\documentclass[10pt,twocolumn,letterpaper]{article}

\usepackage[pagenumbers]{cvpr}      % To produce the REVIEW version
\usepackage[dvipsnames]{xcolor}

\usepackage{url}

\usepackage{algorithm}
\usepackage{algorithmic}

\usepackage[utf8]{inputenc} % allow utf-8 input
\usepackage[T1]{fontenc}    % use 8-bit T1 fonts
\usepackage{multirow}
\usepackage{tikz}

\usepackage{booktabs}
\usepackage[font=small]{caption}

\usepackage{url}            % simple URL typesetting
\usepackage{booktabs}       % professional-quality tables
\usepackage{amsfonts}       % blackboard math symbols
\usepackage{nicefrac}       % compact symbols for 1/2, etc.
\usepackage{microtype}      % microtypography
\usepackage{xcolor}         % colors
\usepackage{amsmath}
\usepackage{pifont}

\newcommand{\figref}[1]{Fig.~\ref{#1}}
\newcommand{\tabref}[1]{Table~\ref{#1}}

\newcommand{\eqnref}[1]{Eqn.~\ref{#1}}

\definecolor{cvprblue}{rgb}{0.21,0.49,0.74}
\usepackage[pagebackref,breaklinks,colorlinks,citecolor=cvprblue]{hyperref}

\def\paperID{3808} % *** Enter the Paper ID here
\def\confName{CVPR}
\def\confYear{2024}

\title{Test-Time Adaptation for Depth Completion}

\author{Hyoungseob Park\\
Yale Vision Lab\\
{\tt\small hyoungseob.park@yale.edu}
\and
Anjali Gupta \\
Yale Vision Lab\\
{\tt\small anjali.gupta@yale.edu}
\and
Alex Wong \\
Yale Vision Lab\\
{\tt\small alex.wong@yale.edu}
}

\begin{document}
\maketitle

\begin{abstract}
It is common to observe performance degradation when transferring models trained on some (source) datasets to target testing data due to a domain gap between them.
Existing methods for bridging this gap, such as domain adaptation (DA), may require the source data on which the model was trained (often not available), while others, i.e., source-free DA, require many passes through the testing data.
We propose an online test-time adaptation method for depth completion, the task of inferring a dense depth map from a single image and associated sparse depth map, that closes the performance gap in a single pass.
We first present a study on how the domain shift in each data modality affects model performance.
Based on our observations that the sparse depth modality exhibits a much smaller covariate shift than the image, we design an embedding module trained in the source domain that preserves a mapping from features encoding only sparse depth to those encoding image and sparse depth. During test time, sparse depth features are projected using this map as a proxy for source domain features and are used as guidance to train a set of auxiliary parameters (i.e., adaptation layer) to align image and sparse depth features from the target test domain to that of the source domain. We evaluate our method on indoor and outdoor scenarios and show that it improves over baselines by an average of 21.1\%. 
Code available at \href{https://github.com/seobbro/TTA-depth-completion}{github.com/seobbro/TTA-depth-completion}.
\end{abstract}

\vspace{-0.0mm}

\section{Introduction}

Reconstructing the 3-dimensional (3D) structure of an environment can support a number of spatial tasks, from robotic navigation and manipulation to augmented and virtual reality. Most systems addressing these tasks are built for sensor platforms equipped with range (i.e., lidar or radar) or optics (i.e., camera or sensors). While range sensors can measure the 3D coordinates of the surrounding space, they often yield point clouds that are sparse. Likewise, these coordinates can also be estimated from images by means of Structure-from-Motion (SfM) or Visual Inertial Odometry (VIO). For the goal of dense mapping, depth completion is the task of recovering the dense depth of a 3D scene as observed from a sparse point cloud, which is often post-processed into a sparse depth map by projecting the points onto the image plane, and guided by a synchronized calibrated image. 

Training a depth completion model can be done in a supervised (using ground truth) or unsupervised (using SfM) manner. The former dominates in performance, but requires expensive annotations that are often unavailable; the latter uses unannotated images, but they must satisfy SfM assumptions between frames, i.e., motion, covisibility, etc.
Like most learning-based methods, models trained under both paradigms typically experience a performance drop when tested on a new dataset due to a covariate shift, i.e., domain gap.
As we can only assume that a single pair of image and sparse depth map is available in the target domain for the depth completion, models belonging to either learning paradigms cannot easily be trained or adapted to the new domain even when given the testing data.
We focus on test-time adaptation (TTA) for depth completion, where one is given access to the test data in a stream, i.e., one batch at a time, without being able to revisit previously-seen examples.
The goal is to learn causally and to quickly adapt a set of pre-existing weights trained on a source domain to a target test domain, so one can reduce the performance gap.

\begin{figure*}
    \centering
    \includegraphics[width=0.93\linewidth]{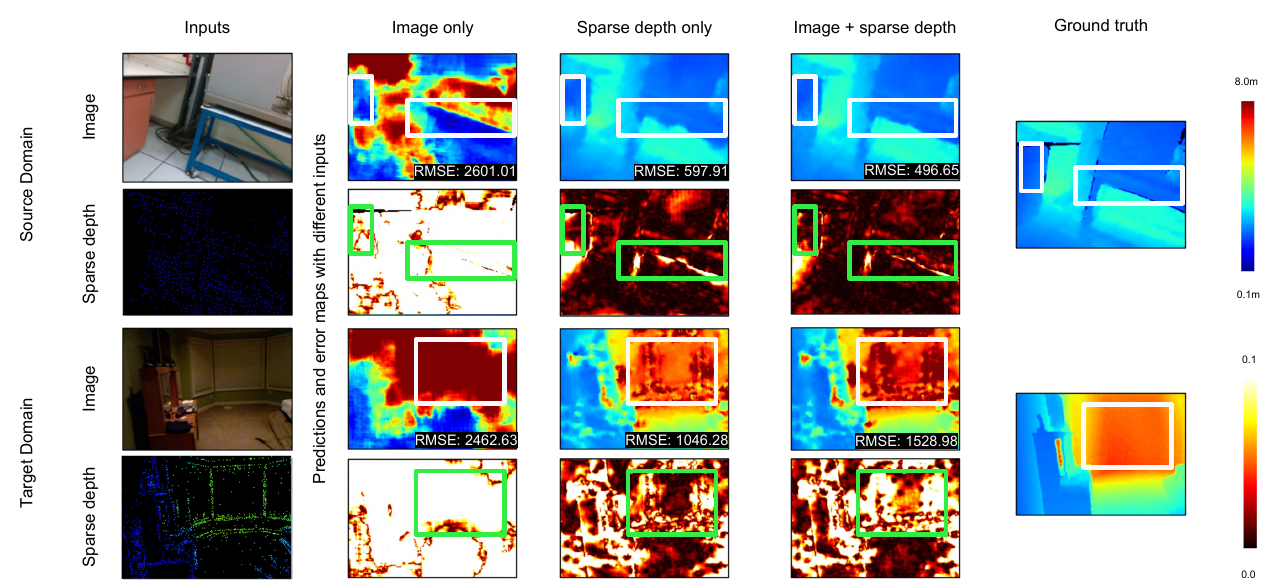}
    \vspace{-2.5mm}
    \caption{
    \textit{Model sensitivity to input modalities.} While utilizing both sparse depth and image as input, the best performance is achieved in the source domain (VOID). Yet, forgoing the image in the test domain (NYUv2) often yields lower error than using both as input.
    %  Columns 2-4: Predictions and error maps of a pretrained model on source (VOID) and target domains using different inputs.
    }
    \label{fig:observations}
    \vspace{-3.6mm}
\end{figure*}

We begin with some motivating observations on the effects of the domain gap: (i) Errors in target domain tend to be higher when feed both the image and sparse depth as input rather than \textit{sparse depth only}, as shown in \figref{fig:observations}.
This implies that the depth modality exhibits a % remarkably
smaller covariate shift between the source and target domains than the image modality, to the extent that forgoing the image altogether often yields superior results than using either both sparse depth and image or the image alone. (ii) Yet, 
when operating
in the source domain, we observe the opposite effect -- forgoing the image is detrimental to 
% the model 
performance. Naturally, this begs the question: How should one leverage data modalities that are less sensitive to the domain shift (e.g., sparse depth) to support alignment between source and target domains for modalities that are more sensitive (e.g., RGB image)? 

To answer this question, we investigate a test-time adaptation approach that learns an embedding for guiding the model parameter update by exploiting the data modality (sparse depth) that is less sensitive to the domain shift. The embedding module maps the latent features encoding sparse depth to the latent features encoding both image and sparse depth.
The mapping is trained in the source domain and frozen when deployed to the target domain for adaptation. During test time, sparse depth is first fed through the encoder and mapped, through the embedding module, to yield a proxy for image and sparse depth embeddings from the source domain -- we refer to the embedded sparse depth features as \textit{proxy embeddings}.
Note: As the mapping is learned in the source domain, the proxy embeddings will also follow the distribution of source image and sparse depth embeddings.
Next, both image and sparse depth from the target test domain are fed as input to the encoder. By maximizing the similarity between test-time input embeddings and the proxy embeddings, we align the target distribution to that of the source to reduce the domain gap. In other words, our method exploits a proxy modality for guiding test-time adaptation and we call the approach, \textit{ProxyTTA}. 
When used in conjunction with typical loss functions to penalize discrepancies between predictions and input sparse depth, and abrupt depth transitions, i.e., Total Variation, the embeddings serve as regularization to guide the model parameter update and prevent excessive drift from those trained on the source data.

Following test-time adaptation conventions, we assume limited computational resources, and that inputs arrive in a stream of small batches and must be processed within a time budget without access to the past data. To ensure fast model updates under these constraints, we deploy auxiliary parameters, or an adaptation layer, to be updated while freezing the rest of the network -- thus achieving low-cost adaptation.
We demonstrate our method in both indoor (VIO) and outdoor (lidar) settings across six datasets, where we not only target typical adaptation scenarios where the shift exists between real and synthetic data domains with similar scenes, i.e. from KITTI \cite{uhrig2017sparsity} to Virtual KITTI \cite{gaidon2016virtual}, but also between different scene layouts, i.e., from VOID \cite{wong2020unsupervised} to NYUv2 \cite{Silberman:ECCV12} and SceneNet \cite{mccormac2016scenenet}. Our % proposed 
proxy embeddings consistently improve over baselines by an average of 21.09\% across all methods and datasets. To the best of our knowledge, we are the first to introduce test-time adaptation for depth completion.

\section{Related work}

\textbf{Test Time Adaptation} (TTA) aims to adapt a given model, pretrained on source data, to test data without access to the source training data. Related fields along this vein include unsupervised domain adaptation~\cite{ganin2015unsupervised, peng2019moment}, which utilizes  source domain data (in practice, this may not be available) for adaptation, and source-free domain adaptation~\cite{kim2021domain}, which does not assume access to source data, but allows access to test data on multiple passes.
In contrast, we focus on test-time adaptation where we 
do not have access to source data and must adapt to test data in a single pass.

Previous studies have proposed strategies to select the source model's component to be preserved, such as the class prototypes extracted from the source data~\cite{choi2022improving, liu2021ttt++, shin2022mm}, the subset of source model parameters~\cite{iwasawa2021test,wang2020tent}, and the discriminative feature from the self-supervised learning (SSL) \cite{choi2022improving}.
For instance, \cite{wang2020tent} proposes TENT, a simple but effective batch-norm layer adaptation with entropy minimization for fully test-time adaptation.
TTT \cite{sun2020test} performs classification-layer adaptation by updating the last linear layer of the source model; \cite{liu2021ttt++} extends this with TTT++ and utilizes joint task-specific and model-specific information based on self-supervised learning.
\cite{iwasawa2021test} presents T3A, an optimization-free classifier adjustment module.
\cite{choi2022improving} uses shift-agnostic weight regularization (SWR) to prevent an effect from the erroneous signal in test time, jointly with the nearest source prototype classifier and a self-supervised proxy task.
\cite{chen2022contrastive} proposes a contrastive learning %-based framework 
with an online pseudo-label refinement while \cite{shin2022mm} proposes % a fully test-time adaptation with 
pseudo-label refinement and momentum update for 3D point cloud segmentation. 
\cite{wang2022continual} proposes continual test-time adaptation based on stochastic restoration and weight-averaged pseudo-labels.
\cite{song2023ecotta} uses efficient residual modules to realign the pretrained weights.
The above methods % are predominantly for 
largely focus on single-image-based tasks, i.e., classification \cite{chen2022contrastive,choi2022improving,liu2021ttt++} and semantic segmentation~\cite{shin2022mm}, and rely on entropy constraints from \cite{wang2020tent}.

Unlike prior work on classification \cite{liu2021ttt++,sun2020test} and segmentation \cite{shin2022mm}, depth completion is a regression problem; hence, existing methods using entropy-based objectives \cite{wang2020tent}, which operate on logits, are not applicable in this task. 
Instead, we propose to minimize sparse depth reconstruction and local smoothness objectives -- similar to that of some existing unsupervised methods \cite{wong2020unsupervised,wong2021unsupervised} -- and to maximize cosine similarity between the proxy embeddings and the test time image and sparse depth embeddings.

\textbf{Depth Completion} aims to output dense depth from a single image and synchronized point cloud, i.e., from lidar or tracked by VIO, projected onto a sparse depth map, by multimodal fusion~\cite{upadhyay2023enhancing,wong2020unsupervised, yang2022touch,yang2023generating,yang2024unitouch,dou2024tactile}.

\textit{Unsupervised depth completion} approaches rely on Structure-from-Motion (SfM) for training and require access to an auxiliary training dataset containing stereo image pairs \cite{shivakumar2019dfusenet,wong2019bilateral} or monocular videos \cite{ma2019self,wong2021learning,wong2021adaptive,wong2020unsupervised,wu2023augundo,liu2022monitored} with synchronized sparse depth maps.
Typically, they minimize a linear combination of photometric reprojection consistency, sparse depth reconstruction error, and local smoothness \cite{ma2019self,wong2021learning,wong2021adaptive,wong2020unsupervised}. These methods can support online training, but are limited by the need for stereo or monocular videos with sufficient parallax and co-visibility. In contrast, our approach does not rely on SfM and can be used in more general scenarios.

\textit{Supervised depth completion} trains the model by minimizing a loss with respect to ground truth. \cite{chen2019learning,huang2019hms} focus on network operations and designs to effectively deal with sparse inputs. \cite{jaritz2018sparse,ma2019self,yang2019dense} propose early and late fusion of image and depth encoder features while \cite{hu2021penet} uses separate networks for each.
\cite{li2020multi} proposes a multi-scale cascaded hourglass network to enhance the depth encoder with image features. \cite{cheng2020cspn++} proposes convolutional spatial propagation network;
\cite{park2020non} extends it to non-local spatial propagation to refine an initial depth map based on confidence and learnable affinity; \cite{lin2022dynamic} further extends it to dynamic spatial propagation.
\cite{eldesokey2020uncertainty,eldesokey2018propagating,qu2021bayesian,qu2020depth} learn uncertainty of the depth estimates. \cite{van2019sparse} utilizes confidence maps to combine depth predictions % from the multi-modal inputs.
while \cite{qiu2019deeplidar,xu2019depth,zhang2018deep} use the surface normals to guide depth prediction.
\cite{kam2022costdcnet} incorporates cost volume for depth prediction. \cite{singh2023depth} used radar.
\cite{rho2022guideformer} devises transformer architecture with cross-modal attention, and \cite{youmin2023completionformer} proposes a hybrid convolution-transformer architecture for depth completion.

While both unsupervised and supervised methods have demonstrated strong performance on benchmarks, they often fail to generalize to test datasets with large domain discrepancies. Moreover, obtaining ground truth is unrealistic for real-time applications, and accumulating sufficient parallax incurs large latencies -- presenting significant challenges for online adaptation. Unlike past works,
we do not assume access to ground truth nor data outside of the input.

\textbf{Unsupervised Domain Adaptation} (UDA) addresses the discrepancy between labeled source data and unlabeled target data \cite{ganin2015unsupervised,peng2019moment,sun2019unsupervised,cicek2019unsupervised}.
The only existing UDA depth completion method \cite{lopez2020project} models the domain gap as the noise in sparse points and the appearance in images.
Unlike most UDA approaches that require source data during adaptation, we are only given the inputs necessary for inference in a stream without the ability to revisit past data, and must update the model online under a limited computational budget.

\textbf{Our Contributions.} We present (i) a study on how the domain shift in each data modality (e.g., image and sparse depth) affects model performance when transferring it from source to target test domain. This study motivates (ii) our approach to learn an embedding of sparse depth features (which are less sensitive to the domain shift) that serves as proxy to source features for guiding test-time adaptation. (iii) To the best of our knowledge, we are the first to propose test-time adaptation for the depth completion task, and (iv) will release code, models, and dataset benchmarking setup to make development accessible for the research community.

\begin{figure*}
\vspace{-5mm}
    \centering
    \includegraphics[width=0.88\linewidth]{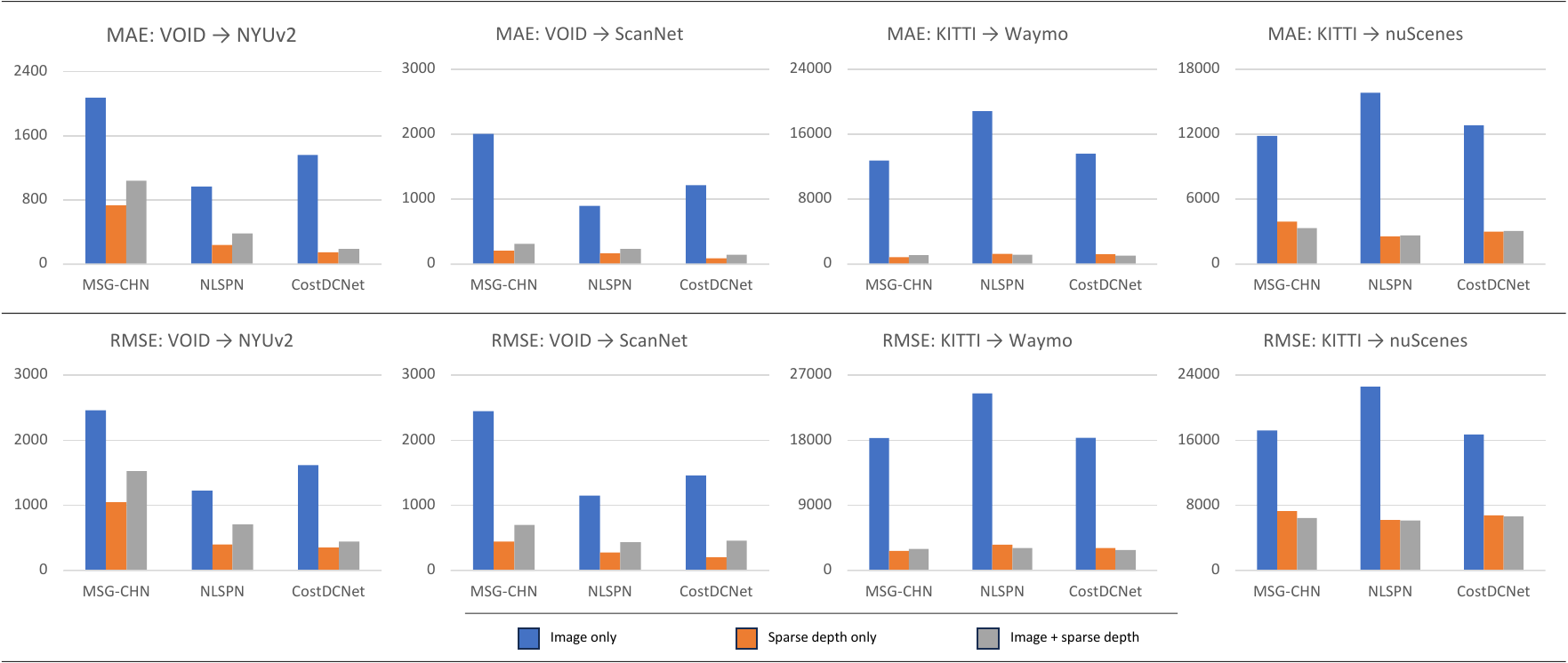}
    \vspace{-2.5mm}
    \caption{\textit{Model sensitivity to input modalities.}  % Using just sparse depth improves over both modalities in test domain.
Depth completion networks have a high reliance on sparse depth modality. Performing inference in a novel domain without the RGB image, i.e., using just sparse depth as input, can improve over using both data modalities.
    }
    \label{fig:preliminary_global}
    \vspace{-3mm}
\end{figure*}

\section{Method Formulation}
For ease of use, we assume access to a % publicly available and % off-the-shelf 
(source) pretrained depth completion model $f_\theta$ that infers a dense depth map $\hat{d}$ from a calibrated RGB image $I \in \mathbb{R}^{H \times W \times 3}$ and its associated sparse point cloud projected onto the image plane as a sparse depth map $z \in \mathbb{R}_+^{H \times W}$, i.e., $f_\theta(I, z) \rightarrow \hat{d} \in \mathbb{R}_{+}^{H \times W}$. For simplicity, we assume that the model was trained to minimize a supervised loss between  prediction and ground truth $d \in \mathbb{R}_+^{H \times W}$ on a source dataset $\mathcal{D}_s = \{I_s^{(n)}, z_s^{(n)}, d^{(n)}\}_{n=1}^{N_s}$, where $N_s$ indicates the number of data samples. Following conventions in TTA, we assume access to the source domain dataset prior to deployment. % Once deployed, we no longer have access to the source data. 

% Likewise, prior to deploying to the target test domain, we assume the same access to supervision on the source dataset. 

During test-time adaptation, we follow the protocol of \cite{liu2021ttt++,sun2020test}, where % the pretrained model 
we only have access to the target domain data $\mathcal{D}_t = \{I_t^{(n)}, z_t^{(n)}\}_{n=1}^{N_t}$ and utilize an online procedure % (i.e. deploying auxiliary parameters) 
to adapt to unseen $N_t$ target data samples. Note that we make no assumptions about supervision during test-time; hence, while we present results on supervised methods for controlled experiments, we see our method being applicable towards unsupervised methods as well.

\begin{figure*}[t]
  \centering
  \includegraphics[width=0.99\linewidth]{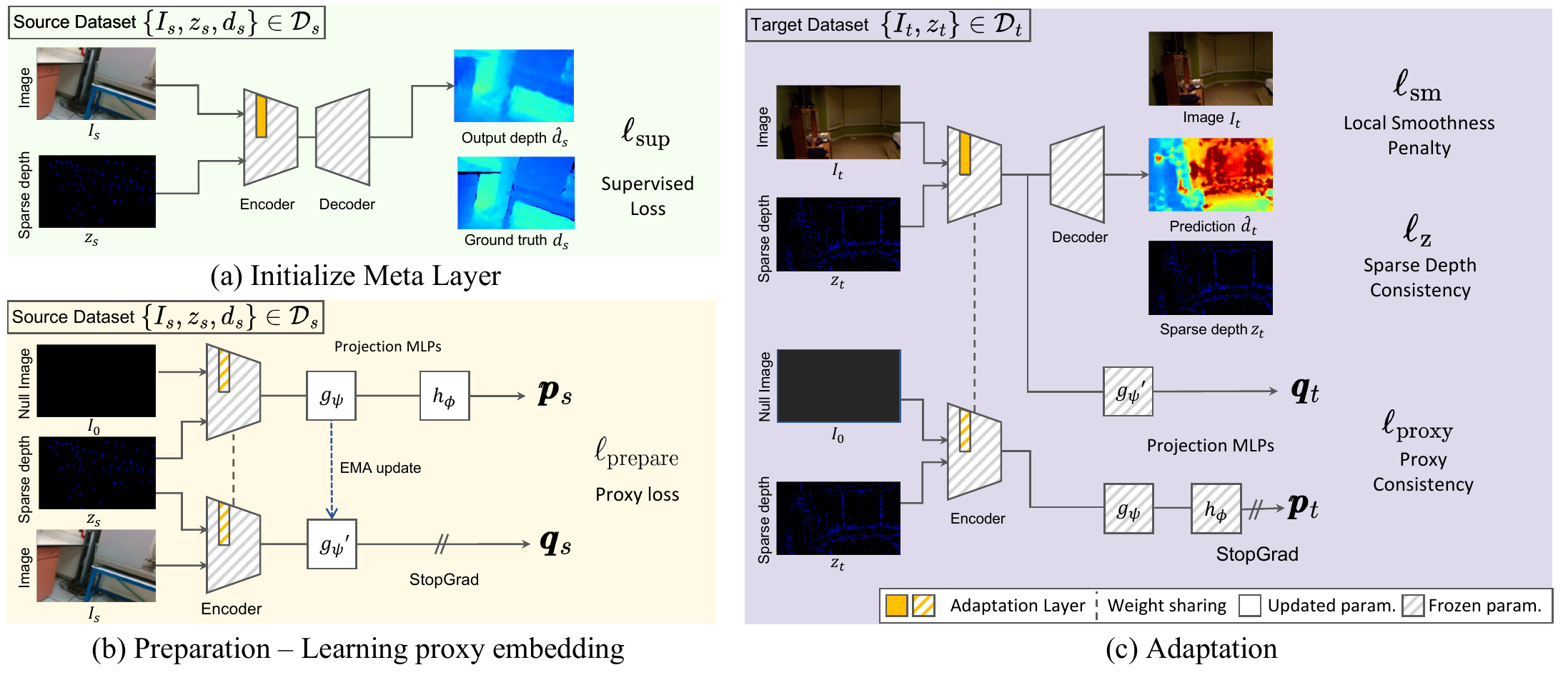}
  \vspace{-3mm}
  \caption{\textit{Overview}. (a) The pretraining stage integrates an adaptation layer into a pretrained encoder and pretrains the adaptation layer on the source dataset. (b) The preparation stage learns the proxy mapping of features encoding sparse depth to those encoding both inputs. (c) The adaptation stage deploys the model to the target domain and updates the adaptation layer by leveraging proxy embeddings as guidance.}
  \label{fig:method_overview}
  \vspace{-2mm}
\end{figure*}

Our method, ProxyTTA, is split into three stages (\figref{fig:method_overview}): (a) During an intialization stage, we augment the network encoder with an adaptation layer and train it using source domain data. (b) In the preparation stage, we learn a mapping from sparse depth features to image and sparse depth (proxy) embeddings. (c) During test time, we do not need the source dataset; we freeze the mapping and use its proxy embeddings for updating the adaptation layer parameters in test domain.

\subsection{Sensitivity Study on Data Modalities}
\label{sec:sensitivity}
To motivate our approach, we begin with a sensitivity study of depth completion networks to input modalities, e.g. image, sparse depth, and the effect of domain shift on them. To this end, we alter the inputs by zeroing out either $I$ or $z$ to yield $(I,z), (I_0, z)$, and $(I, z_0)$, where $I_0$ and $z_0$ indicate the zero matrices with identical size to $I$ and $z$, respectively. We evaluate the pretrained models using $(I,z), (I_0, z)$, and $(I, z_0)$ to highlight their dependence on each input modality and to gauge their sensitivity when one modality gives no useful information at all. 
\figref{fig:observations} and \figref{fig:preliminary_global} show qualitative (error maps) and quantitative results (bar graphs), respectively, of pretrained depth completion models when fed the different inputs on the source dataset $\mathcal{D}_s$ and the target dataset $\mathcal{D}_t$.

In the source domain, inference using both image and sparse depth as inputs, i.e., $\hat{d}_{s}(I, z)$, shows the best performance. Surprisingly, the inference using sparse depth alone (i.e., with null-image) $\hat{d}_{s}(I_0, z)$ is comparable to $\hat{d}_{s}(I, z)$. This shows the first intuition behind our approach: (i) Even though depth inputs are sparse, they are sufficient to support the reconstruction of the scene. Additionally, inference with image alone (i.e., null-depth) $\hat{d}_{s}(I, z_0)$ is worse than $\hat{d}_{s}(I_0, z)$ and $\hat{d}_{s}(I, z)$, which suggests that a depth completion network relies heavily on sparse depth modality for inference, and the image for guiding recovery of finer details. % in completion.

In the target test domain, expectedly, performance degrades for inference using both image and sparse depth due to a covariate shift. Remarkably, we observe that predictions from sparse depth alone $\hat{d}_{t}(I_0,z)$ remain consistent in performance to those using both inputs $\hat{d}_{t}(I,z)$. Moreover, we observe that in most cases $\hat{d}_{t}(I_0,z)$, in fact, outperforms $\hat{d}_{t}(I,z)$ across several methods and datasets, i.e., inference without image information in the test domain is better than with it. Conversely, the performance gap between inference with both inputs, $\hat{d}_{t}(I,z)$, and just the image, $\hat{d}_{t}(I,z_0)$, becomes more evident under the domain shift. This observation illustrates another intuition: (ii) The domain shift largely affects the image modality, and less so depth.

The two intuitions above motivate our approach.
As object shapes tend to persist across domains, and the measured sparse points being a coarse representation of them, we aim to leverage sparse depth modality to bridge the domain gap.
To this end, we exploit the observation that depth completion networks are able to recover (coarse) 3D scenes from sparse points alone and that the image serves to propagate and refine depth for regions lacking points.
This is done by learning to map features encoding sparse depth inputs to features encoding both modalities in the source domain and, during test-time, recover the source domain features compatible with target domain sparse depth to guide model adaptation.
Specifically, as observed, the covariate shift is largely photometric, so we propose to adapt the RGB image encoder branch by introducing a \textit{adaptation layer}: a single convolutional layer designed to align target domain RGB embeddings to those of the source domain.
As the rest of the network is frozen, adapting just the adaptation layer allows for low-cost model updates. 

\textbf{Intuition for integrating adaptation layer.} Guided by our observations, the adaptation layer should be (i) placed in the image encoder branch prior to the fusion of image and depth features, and (ii) located within later layers to modulate higher level representations (i.e., object shapes, as opposed to low-level edges). (iii) connected as a skip connection to decoder to more directly affect the output.

\subsection{Preparation Stage - Source Domain}
\label{preparation}
\textbf{Initialize adaptation layer from source domain.}
Updating the entire network is largely infeasible in test-time adaptation scenarios. For the sake of speed and efficiency, we implement an adaptation layer $m_\phi$, i.e., a convolutional layer, within the encoder of a pretrained network. Note that the entire network will be frozen during all stages of our method with the exception of the adaptation layer and proxy mapping, where both will be initialized during preparation stage in the source domain; the proxy mapping will then be frozen and used to adapt $m_\phi$ in the target domain. To ease the adaptation process, we initialize the $m_\phi$ by minimizing a supervised loss over the source dataset~(\figref{fig:method_overview}-(a)). We denote the pretrained encoder integrated with $m_\phi$ as $e_\phi$.

\noindent\textbf{Learning proxy mapping from source domain.} As observed in \figref{fig:observations}, the best results in the source domain are achieved by feeding in both the image and sparse depth modalities for inference. 
However, the image is susceptible to domain shift which degrades performance when the model is transferred to an unseen test domain. Conversely, sparse depth is more resilient to the domain shift than RGB images, i.e., the shape of a car (or another object) remains similar regardless of (synthetic or real) domain.
Our method aims to leverage the sparse depth modality, which is less sensitive to the domain shift, in the downstream adaptation process.
To this end, we employ a soft mapping \cite{grill2020bootstrap} from just the encoded sparse depth features to sparse depth \textit{and image features} to learn the photometric information that is captured from the same scene as the sparse point cloud.
This strategy allows us to learn the mapping that projects the sparse depth features to ``proxy'' embeddings close to those that also encode the image. In other words, it fills in what is missing in the image encoder branch by predicting the residual latent image encoding that is compatible with the input sparse depth, i.e., 3D scene. As this is trained in the source domain, the mapping naturally yields proxy embeddings that encode the source domain image (and sparse depth), which can be later used to guide the adaptation layer $m_\phi$ to transform test domain RGB features close to those of the source domain.

This mapping by MLPs can be denoted as $g_\psi(\cdot), g_{\psi'}(\cdot)$ and $h_\omega(\cdot)$; to learn them, we get two embeddings $\pmb{p}_s$ and $\pmb{q}_s$,
\begin{equation}
    \vspace{-1mm}
    \begin{aligned}
        \pmb{p}_s = h_\omega(g_\psi(\text{StopGrad}(e_\phi({I_0}, z_s)))), \\ % \,\,\, 
        \pmb{q}_s = \text{StopGrad}({g_{\psi'}}(e_\phi(I_s, z_s)))
    \end{aligned}
    \vspace{-0.5mm}
\end{equation}

% where $e_\phi$ denotes the encoder from the source pretrained network, including the adaptation layer, and {\o}  indicates a zero matrix with the same size as image inputs.
where $e_\phi$ denotes the encoder augmented with the adaptation layer trained on source dataset, $I_s, z_s$, the image and sparse depth from source domain, and $I_0$ the null-image. The embedding modules $g_\psi$ and $h_\omega$ are updated to maximize the similarity between $\pmb{p}_s$ and $\pmb{q}_s$.
To learn them, we minimize:
\begin{equation}
\vspace{-1mm}
    \ell_\text{prepare} = 1 - (\frac{\pmb{p}_s}{\|\pmb{p}_s\|} \cdot \frac{\pmb{q}_s}{\|\pmb{q}_s\|}),
\label{eq:cosine}
\vspace{-0.5mm}
\end{equation}
where $\|\cdot\|$ is $L2$-norm, and $(\pmb{a}\cdot \pmb{b})$ indicates the dot product of the vectors $\pmb{a}$ and $\pmb{b}$.
To this end, we first train the MLP heads $g_\psi, h_\omega$ by minimizing \eqnref{eq:cosine}. Note that the MLP head $g_{\psi'}$ is updated with EMA update following BYOL~\cite{grill2020bootstrap} to avoid collapse: $g_{\psi'} \leftarrow \tau\cdot g_{\psi'} + (1-\tau)\cdot g_{\psi}$.

Once the mapping is learned, we can freeze the embedding module and deploy it for test-time adaptation where we update the adaptation layer weights $\phi$ to maximize the similarity between the embeddings of a test domain image and sparse depth, and its proxy from the source domain.
Naturally, due to the domain shift, the embeddings will yield low similarity scores; hence, maximizing the scores through our proxy embedding implicitly aligns the target RGB distribution to that of the source distribution, i.e., minimizing the cosine similarity between the source and target distributions.

\subsection{Deploying Proxy Mapping to Target Domain}
\label{adaptation}
%- in conclusion sparse depth feature serves as an anchor feature

\textbf{Adaptation stage} aims to update the adaptation layer parameters by minimizing a test-time loss function over the target test domain data $\{I_t, z_t\} \in \mathcal{D}_t$. To do so, we deploy the learned proxy mapping module (MLP heads $\{g^*_\psi(\cdot),g^*_{\psi'}(\cdot),$ and $ h^*_\omega(\cdot)\}$) along with the adaptation layer $m_\phi$ integrated into the frozen encoder as $e_\phi$.

\textbf{Adaptation loss.} For adaptation, our loss is composed of a linear combination of three loss terms:
\begin{equation}
    \mathcal{L}_{\text{adapt}} = w_z \ell_z + w_{\text{sm}} \ell_{\text{sm}} + w_{\text{proxy}} \ell_{\text{proxy}}, 
    \vspace{-1mm}
    \label{eqn:loss_adapt}
\end{equation}
where $\ell_z$, $\ell_\text{sm}$ denote sparse depth consistency loss and local smoothness loss, respectively,  $\ell_{\text{proxy}}$ is proxy mapping consistency loss, and $w$ indicates a weight of each loss term.

\textbf{Sparse Depth Consistency.} 
Sparse point clouds capture a coarse structure of the 3D scene.To obtain metric scale predictions consistent with the scene structure, we minimize L1 error between the sparse depth $z_t$ and the prediction $\hat{d}_t$: % over the sparse depth map domain.
\begin{equation}
    \ell_z = \frac{1}{|\Omega{(z_t)|}} \sum_{x \in \Omega{(z_t)}} |\hat{d}_t(x) - z_t(x)|,
    \label{eqn:loss_sparse_depth}
    \vspace{-1mm}
\end{equation}
where $x \in \Omega{(z_t)}$ are the pixel locations where sparse points were projected onto the image plane.

\textbf{Local Smoothness.} % As depth estimation is an ill-posed inverse problem, we need regularization. 
Based on the assumption of local smoothness and connectivity in a 3D scene, we impose the same in the predicted depth map $\hat{d}_t$. Specifically, we apply an L1 penalty to its gradients in both the x- and y-directions (i.e., $\partial_X$ and $\partial_Y$). We balance the weight of each term with $\lambda_X$ and $\lambda_Y$, to allow discontinuities over object boundaries based on the image gradients, where $\lambda_X(x) = e^{-|\partial_X I_t(x)|}$, $\lambda_Y(x) = e^{-|\partial_Y I_t(x)|}$, and $\Omega$ denotes the image domain.
% The weight for x- and y-direction penalty terms, $\lambda_X$ and $\lambda_Y$, are denoted as $\lambda_X(x) = e^{-|\partial_X I_t(x)|}$ and $\lambda_Y(x) = e^{-|\partial_Y I_t(x)|}$. 
\begin{equation}
    \ell_{\text{sm}} = \frac{1}{|\Omega|} \sum_{x \in \Omega} \lambda_X(x)|\partial_X \hat{d}_t(x)| + \lambda_Y(x)|\partial_Y \hat{d}_t(x)|.
    \label{eqn:loss_smoothness}
\end{equation}

\begin{table*}[t]
\scriptsize
\centering
\setlength\tabcolsep{3pt}
\resizebox{0.94\textwidth}{!}{%
\begin{tabular}{cl cccc cccc cccc}
    \midrule
    Method & {} & MAE & RMSE & MAE & RMSE & MAE & RMSE \\
    \midrule
    &  &  \multicolumn{2}{c}{KITTI $\rightarrow$ VKITTI-FOG} & \multicolumn{2}{c}{KITTI $\rightarrow$ nuScenes} & \multicolumn{2}{c}{KITTI $\rightarrow$ Waymo}\\
    \midrule 
    \multirow{4}{*}{MSG-CHN} & Pretrained & 2842.88 & 6557.38 & 3331.821 &   6449.094 & 1107.22& 2962.45 \\
    & CoTTA &\textit{730.6$\pm$11.67}& \textit{3330.23$\pm$44.83} & \textit{3157.69} & \textit{6434.14} & \textit{655.77$\pm$30.98}&\textit{2213.27$\pm$98.80}\\ 
    & ProxyTTA-fast (Ours) & \textbf{728.24}$\pm$\textbf{3.73} & \textbf{3087.36}$\pm$\textbf{15.92} & \textbf{2834.08$\pm$17.64}&\textbf{6096.56$\pm$21.08} & \textbf{608.91$\pm$1.74}&\textbf{1921.83$\pm$2.54} \\ 
    \midrule
    \multirow{6}{*}{NLSPN} &  Pretrained & 1309.99 &  7423.48 &  2656.609           & 6146.590 & 1175.83&3078.377  \\ 
     & BN Adapt & 1140.21$\pm$35.89 & 4592.86$\pm$198.21 & 11291.57$\pm$21.32&16670.87$\pm$52.56 & 7283.33$\pm$104.58&9670.36$\pm$250.22 \\
     & BN Adapt, $\ell_z$, $\ell_{sm}$ & 775.20$\pm$5.65 & 3465.05$\pm$32.73 & 2928.51$\pm$75.89&8209.24$\pm$164.31 & \textit{494.94$\pm$3.08}&\textit{1921.17$\pm$338.06}\\
     & CoTTA &767.93$\pm$5.47&3799.88$\pm$17.29 &\textit{2650.45$\pm$15.04} &6242.52$\pm$33.14 & 933.41$\pm$4.31&2763.88$\pm$143.48\\  
     & ProxyTTA-fast & \textit{732.61$\pm$29.57}& \textit{3002.19}$\pm$ \textit{52.29} & 2733.96$\pm$34.32 & \textit{6099.48$\pm$82.32} & 875.01$\pm$15.8&2400.17$\pm$21.44 \\
     & ProxyTTA (Ours) & \textbf{686.91}$\pm$\textbf{22.14}  & \textbf{2666.70$\pm$56.64} 
     & \textbf{2589.25}$\pm$\textbf{59.03}&\textbf{6006.18}$\pm$\textbf{90.66} & \textbf{477.28$\pm$3.32}&\textbf{1598.64$\pm$18.95}\\
     \midrule  
     \multirow{7}{*}{CostDCNet} &  Pretrained & 1042.98 & 6301.60 &  3064.724          & 6630.649 & 1093.79& 2798.25\\ 
     & BN Adapt & 1476.57$\pm$1.38 & 5428.20$\pm$8.15 & 2306.04$\pm$28.86&6391.98$\pm$48.97 & 596.08$\pm$5.55&1877.91$\pm$45.56 \\
     & BN Adapt, $\ell_z$, $\ell_{sm}$ & \textit{729.67$\pm$3.14} & 3413.76$\pm$14.59 & \textit{2288.85$\pm$14.02}&6338.38$\pm$31.31 & \textit{469.97$\pm$2.47}&\textbf{1572.95$\pm$10.63}\\
     & CoTTA &756.32$\pm$3.59&3686.69$\pm$14.75 & 2676.83$\pm$68.92&\textit{6099.49$\pm$66.79} & 689.94$\pm$1.95&2140.23$\pm$16.12\\   
     & ProxyTTA-fast & 756.98$\pm$31.07 & \textit{3091.78$\pm$105.42} & 2595.81$\pm$12.13&6373.01$\pm$7.74 & 606.10$\pm$11.10&1817.79$\pm$19.14\\  
     & ProxyTTA (Ours) & \textbf{512.72}$\pm$\textbf{0.74}&\textbf{2735.01}$\pm$\textbf{3.53} & \textbf{2062.28$\pm$11.24} &\textbf{5509.96$\pm$23.41} & \textbf{466.44$\pm$1.63}&\textit{1580.38$\pm$11.48} \\
    \midrule
    &  & \multicolumn{2}{c}{VOID $\rightarrow$ NYUv2} & \multicolumn{2}{c}{VOID $\rightarrow$ SceneNet} & \multicolumn{2}{c}{VOID $\rightarrow$ ScanNet} \\ 

    \midrule 
    \multirow{3}{*}{MSG-CHN} & Pretrained & 1040.934 & 1528.983 & 281.28 & 645.01 & 687.988 & 1201.747 \\
    & CoTTA & \textit{876.93$\pm$146.95}&\textit{1148.62$\pm$173.53} & \textit{223.19$\pm$14.77}&\textit{498.46$\pm$28.21} & \textit{619.37$\pm$4.14}&\textit{1141.04$\pm$7.35}\\  
    & ProxyTTA-fast (Ours)& \textbf{699.60}$\pm$\textbf{6.00} & \textbf{1120.37}$\pm$\textbf{9.76} & \textbf{192.74}$\pm$\textbf{1.72} & \textbf{424.49}$\pm$\textbf{4.58} & \textbf{302.21$\pm$4.10} & \textbf{480.08$\pm$8.03} \\ 
    \midrule
    \multirow{6}{*}{NLSPN} &  Pretrained & 388.87 &  702.80 & 167.250  & 438.71 & 233.33 & 431.20 \\ 
     & BN Adapt & 250.13$\pm$5.23  & 447.18$\pm$10.32 & 143.61$\pm$6.34 & 385.56$\pm$9.84 &207.00$\pm$0.57&401.41$\pm$2.84  \\
     & BN Adapt, $\ell_z$, $\ell_{sm}$ & \textit{147.55$\pm$1.36}  & \textit{271.10$\pm$2.17} & \textit{120.48$\pm$1.94}&\textit{345.91$\pm$7.14} & \textit{82.76$\pm$0.47}&\textit{181.97$\pm$1.21} \\
     & CoTTA & 390.50$\pm$8.29&704.72$\pm$16.74 & 205.02$\pm$1.79&540.01$\pm$4.08&234.77$\pm$1.52&496.18$\pm$2.75\\ 
     & ProxyTTA-fast&  168.43$\pm$3.46 & 309.48$\pm$6.92   &124.67$\pm$1.33&357.56$\pm$2.59 & 104.06$\pm$11.03&232.84$\pm$20.46   \\
     & ProxyTTA (Ours)& \textbf{124.41}$\pm$\textbf{2.27} & \textbf{240.73}$\pm$\textbf{5.72} &\textbf{113.93$\pm$1.49}&\textbf{333.41$\pm$4.32}& \textbf{74.77$\pm$0.31} & \textbf{166.61$\pm$0.45} \\
     \midrule  
     \multirow{6}{*}{CostDCNet} &  Pretrained & 189.10   & 446.71& 173.37  & 443.22  &144.31 &  458.69 \\ 
     & BN Adapt & 160.31$\pm$2.7&410.55$\pm$10.70&176.62$\pm$0.72&446.32$\pm$8.52&  159.65$\pm$4.63&399.14$\pm$13.92  \\
     & BN Adapt, $\ell_z$, $\ell_{sm}$ &136.80$\pm$5.35&338.59$\pm$22.36&134.22$\pm$2.33&385.9$\pm$6.68 & \textit{68.44$\pm$0.46} &\textit{164.59$\pm$2.82} &   \\
     & CoTTA & 147.69$\pm$5.3&376.87$\pm$21.25 &136.42$\pm$3.41&405.38$\pm$11.63 & 101.98$\pm$1.53&322.63$\pm$5.04 \\  
     & ProxyTTA-fast& \textit{131.93$\pm$2.58}&\textit{269.02$\pm$5.61} &\textit{129.99$\pm$3.88}&\textbf{353.86$\pm$7.91}& 128.12$\pm$3.41&244.62$\pm$7.53  \\
     & ProxyTTA (Ours)&   \textbf{95.87$\pm$2.16}&\textbf{203.83$\pm$4.72} &\textbf{125.75$\pm$1.93}&\textit{357.12$\pm$4.13} &\textbf{68.17$\pm$0.44} &\textbf{162.35$\pm$1.12}   \\
    \midrule
\end{tabular}
    }
\vspace{-1mm}
\caption{\textit{Qualitative results.} For indoors, we adapt from VOID to NYUv2, SceneNet, and ScanNet; for outdoors, from KITTI to VKITTI with fog, nuScenes, and Waymo. \textbf{Bold} denotes best and \textit{Italics} second-best. ProxyTTA-fast denotes our method without updating BatchNorm.% \textbf{Bold} denotes best and \textit{italics} runner-up.
}
\vspace{-3mm}
\label{tab:experiments:outdoor}
\end{table*}

\textbf{Proxy Consistency.}
In order to regularize the adaptation with the learned mapping from the previous stage, we freeze the weight parameters of MLP heads $\{g^*_\psi(\cdot), h^*_\omega(\cdot)\}$, and update the parameters of the adaptation layer $m_\phi$. First, we obtain the features $\pmb{p}_t$ and $\pmb{q}_t$ using the null-image $I_0$ in one and the given target test domain image $I_t$ in the other:
\begin{equation}
    \begin{aligned}
    \pmb{p}_t &= \text{StopGrad}(h^*_\omega(g^*_\psi(e_\phi({I_0}, z_t)))), \,\,\, \\ \pmb{q}_t &= g^*_{\psi'}(e_\phi(I_t, z_t)).
    \end{aligned}
\end{equation}

% where \textit{sg} denotes the stop gradient operation.
We maximize the cosine similarity between the feature $\pmb{q}_t$ and $\pmb{p}_t$ via a proxy loss $\ell_{\text{proxy}}$ to update adaptation layer $m_\phi$:
\begin{equation}
    \ell_{\text{proxy}} = 1 - (\frac{\pmb{p}_t}{\|\pmb{p}_t\|} \cdot \frac{\pmb{q}_t}{\|\pmb{q}_t\|}).
    \label{eqn:loss_proxy}
\end{equation}
% This process can be interpreted as minimizing the KL divergence between the source distribution $P(D_s)$ and the target distribution $P(D_t)$ by updating adaptation layer. See Supp. mat. 
\section{Experiments}
We demonstrate the effectiveness of our approach on a mix of both real and synthetic datasets including indoor SLAM/VIO scenarios  (VOID~\cite{wong2020unsupervised}, NYUv2~\cite{Silberman:ECCV12}, SceneNet~\cite{mccormac2016scenenet}, and ScanNet~\cite{dai2017scannet}) and outdoor driving scenarios using lidar sensor (KITTI~\cite{uhrig2017sparsity}, Virtual KITTI (VKITTI)~\cite{gaidon2016virtual}, nuScenes~\cite{caesar2020nuscenes}, and Waymo Open Dataset~\cite{sun2020scalability}). 
We chose three representative architectures of current depth completion methods to test our method: MSG-CHN \cite{li2020multi} (CNN-based), NLSPN \cite{park2020non} (SPN-based) and CostDCNet \cite{kam2022costdcnet} (cost volume-based). All reported results are averaged over 5 independent trials. We describe implementation details, hyper-parameters used, hardware requirements, evaluation metrics as well as additional experimental results in the Supp. Mat.

\textbf{Main Result.}
We use pretrained models (MSG-CHN, NLSPN, and CostDCNet) from the two source datasets, VOID for indoor, and KITTI for outdoor.
For indoor, we adapt models pretrained on VOID to NYUv2, SceneNet, and ScanNet; for outdoors, we adapt from KITTI to VKITTI (with fog), nuScenes, and Waymo. 
BN Adapt denotes updating the batch statistics (i.e., running mean and variance).
BN Adapt, $\ell_{z}, \ell_{sm}$ is a variation of TENT \cite{wang2020tent} which minimizes \eqnref{eqn:loss_sparse_depth}, \ref{eqn:loss_smoothness} instead of entropy by updating learnable scale factors.
CoTTA denotes replacing proxy loss with L1 consistency loss w.r.t. the pretrained prediction~\cite{wang2022continual}.
ProxyTTA-fast denotes our method without batch norm update, which improves adaptation runtime by 25.32\%.

\begin{figure*}[h!]
    \vspace{-1cm}
    \centering
    \includegraphics[width=0.92\linewidth]{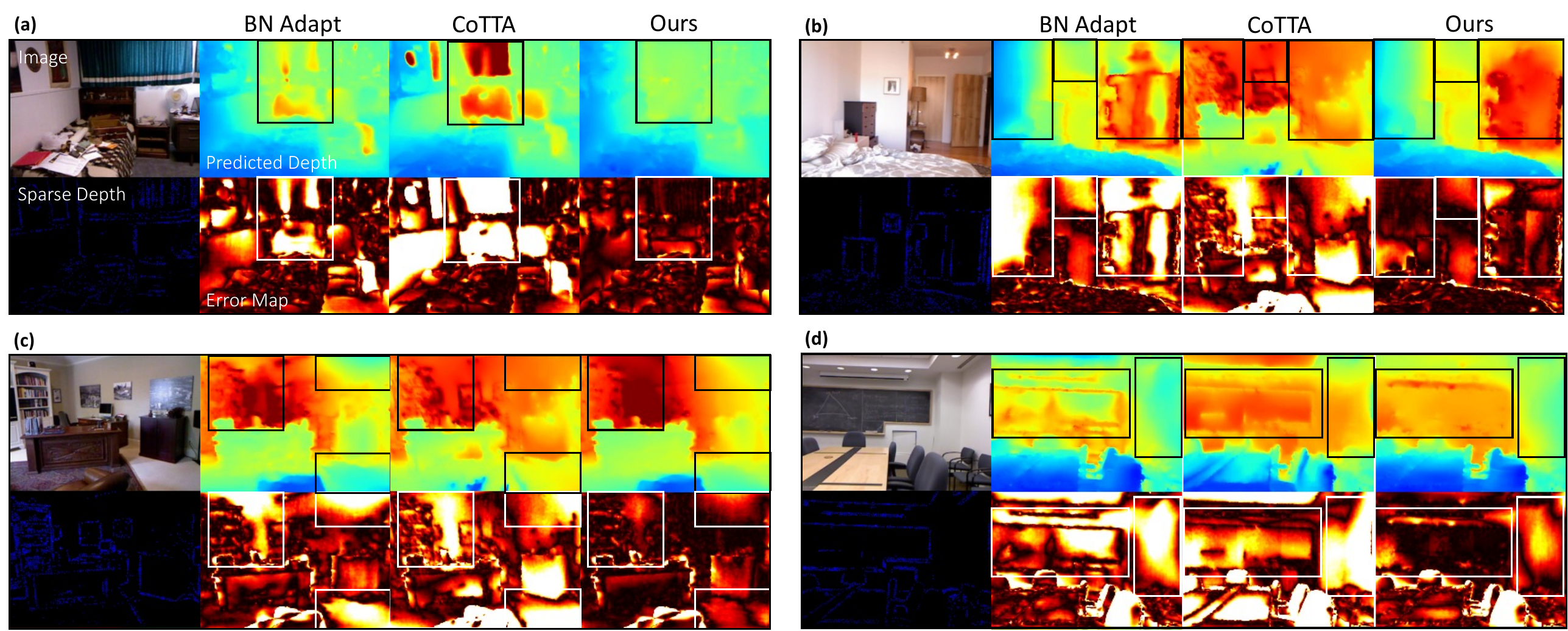}
    \vspace{-0.3cm}
    \caption{\textit{Qualitative results on NYUv2.}
    For indoors scenarios, ProxyTTA performs better in boundary regions displaying the discontinuity in depth (e.g., curtains, (a)), as well as homogeneous regions (e.g., blackboard, (d)). Boxes highlight detailed comparisons. 
    }
    \vspace{-4mm}
    \label{fig:qualitative_indoor}
\end{figure*}

\begin{figure*}[t!]
    \centering
    \includegraphics[width=0.92\linewidth]{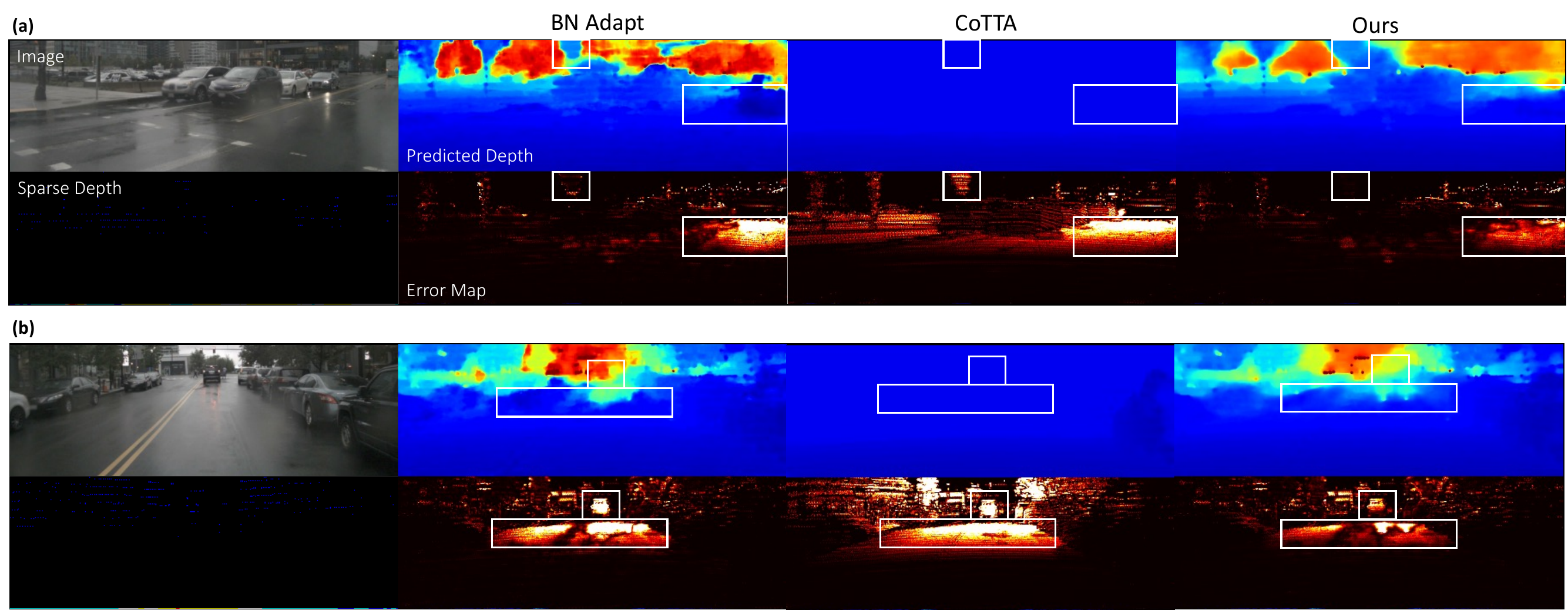}
    \vspace{-0.2cm}
    \caption{\textit{Qualitative results on NuScenes.} For outdoor adaptation scenarios, ProxyTTA improves over BN Adapt and CoTTA, notably in both depth-discontinuous regions (e.g., car in (b)) and homogeneous regions (e.g., road in (a) and (b)). Boxes highlight detailed comparisons.
    }
    \vspace{-2.2mm}
    \label{fig:qualitative_outdoor}
\end{figure*}

Our method consistently improves over baselines and variants of BN Adapt (\tabref{tab:experiments:outdoor}).
Specifically, we improve over BN Adapt, $\ell_{z}, \ell_{sm}$ by 11.60\% on average across all methods for indoor, 19.73\% on outdoors, and 15.67\% overall to achieve state-of-the-art performance.
Qualitatively, ~\figref{fig:qualitative_indoor} and \figref{fig:qualitative_outdoor} show that our method performs better in boundary regions and homogeneous regions, thus exhibiting less oversmoothing on curtains in \figref{fig:qualitative_indoor}-(a) and car in \figref{fig:qualitative_outdoor}-(b), and undersmoothing on blackboard in \figref{fig:qualitative_indoor}-(d) and road in \figref{fig:qualitative_outdoor}-(a), respectively, during adaptation. This trend is due to the proxy loss and the adaptation layer, which allows us to adapt with minimum weight adjustments while preserving high-level features (object shapes) learned from the source domain by mapping the target RGB modality to that of the source domain.
Notably, ProxyTTA-fast still improves over BN Adapt even though we only adapt our adaptation layer, which demonstrates the effectiveness of our design choice as well as our proposed proxy embeddings. We visualize image and sparse depth features from the source and target domains along with proxy embeddings in the target domain using t-SNE \cite{van2008visualizing} in Fig. 1 of Supp. Mat.; we observe that proxy embeddings are close to source domain features.

\textbf{Comparison to BN adaptation\footnote{MSG-CHN lacks Batch Norm (BN) layer so we cannot use BN adapt.} and CoTTA.} 
To assess the impact of our adaptation layer, we compare to batch norm  (BN) adaptation from TENT~\cite{wang2020tent}. 
In BN adaptation, we only update the batch norm layer's scale and shift factor based on the loss function.
On average, BN Adapt with $\ell_{z}, \ell_{sm}$ improves the pretrained model by 32.77\%; whereas, updating just our adaptation layer (ProxyTTA-fast) improves it more by 34.07\% (\tabref{tab:experiments:outdoor}). 
The improvement of ProxyTTA-fast over BN adapt demonstrates the efficacy of updating adaptation layer, which directly adjusts the high-level features from the RGB branch guided by proxy loss, where BN adapt realigns the learned source features from both RGB and range sensors by updating feature statistics.

Nonetheless, the best results are achieved when we include batch norm update, which improves the pretrained model by 44.53\%, but at the cost $\approx$33.2\% of total extra time. The improvement of ProxyTTA over BN adapt implies that the large domain discrepancy may not be addressed by adapting only BN parameters (i.e., scaling and shifting); ProxyTTA explicitly adjusts RGB features by updating the adaptation layer with proxy embeddings as guidance. 

We also compared our approach to CoTTA~\cite{wang2022continual}, which adapts \textit{the whole model parameters} using the prediction from the teacher model updated by exponential moving average of pretrained weight and the model prediction.
We combined additional loss $\ell_\text{z}, \ell_\text{sm}$ on top of CoTTA loss, since we observed that the models cannot be adapted with CoTTA alone.
Specifically, our method without proxy shows a 25.26\% average improvement on the CoTTA method. 
CoTTA updates the whole parameters including RGB and sparse depth branch, which causes a drift from the learned model parameters. On the other side, our method only updates additional layer at RGB branch, based on the study from the most domain discrepancy comes from RGB modality as studied in Sec. \ref{sec:sensitivity}, and this prevents the model from a drift from learned domain. Also, CoTTA assumes the test-time augmentation can mitigate the domain shift. However, the results shows test-time augmentation on RGB image, causing a small distributional shift, may not solve a large domain discrepancy.

Also, our method with batch normalization layer update shows 26.52\% average improvement, while using 25.05\% less adaptation time.
Note: CoTTA costs not only \textit{additional memory for teacher model} but also \textit{inference time to get the teacher model prediction}, even if CoTTA does not require any preparation process.
Overall, our method shows 21.09\% average improvement over BN adapt and CoTTA methods.

% \vspace{-0.5mm}
\section{Discussion}
\vspace{-1.5mm}
We have proposed a method for test-time adaptation for depth completion that leverages the strength of complementary multi-sensor setup in the presence of domain shift.
By studying model sensitivity to each input modality as well as the data under domain shift, we designed a way to exploit the modality (sparse depth) that is less sensitive to guide adaptation. We do so through a proxy embedding that learns the photometric information from the source domain that is compatible with the sparse depth depicting a 3D scene.
Our proxy embedding works well as a regularizer for scenarios where there exists covariate shifts in photometry (i.e., KITTI to VKITTI) as well as scene layouts (i.e., VOID to NYUv2 and SceneNet).
While one may surmise that the application of the embeddings are specific to scene distributions, we show otherwise. VOID (classrooms, laboratories, and gardens), NYUv2 (households and shopping centers), and SceneNet (randomly arranged synthetic rooms) all differ in layouts. The proxy embedding captures latent photometric features of the object shapes populating them; the same proxy embedding can be transferred across domains even when scene differ, but share objects within them.

This leads to possible limitations in the scenarios where the source dataset is sampled from scenes that do not share any objects with the target test dataset; in this case, the proxy embeddings should give little to no gain and one must rely on generic regularizers like local smoothness. Additionally, while we follow the conventions in TTA and assume access to the source dataset prior to deployment, in reality, many models are trained on private datasets, so adapting ``off-the-shelf'' models remains a challenge. In such cases, one must incorporate our preparation pipeline into their model training and release the adaptation layer and proxy embedding module together with network weights. Nonetheless, this is the first test-time adaptation work in depth completion; in addition to our findings, we release models, dataset, adaptation, and evaluation code, and hope to further motivate interest in TTA for multi-modal tasks like depth completion.

\noindent\textbf{Acknowledgements.} This work was supported by NSF 2112562 Athena AI Institute.

{
    \small
    \bibliographystyle{ieeenat_fullname}
    \bibliography{main}

\begin{thebibliography}{63}
\providecommand{\natexlab}[1]{#1}
\providecommand{\url}[1]{\texttt{#1}}
\expandafter\ifx\csname urlstyle\endcsname\relax
  \providecommand{\doi}[1]{doi: #1}\else
  \providecommand{\doi}{doi: \begingroup \urlstyle{rm}\Url}\fi

\bibitem[Caesar et~al.(2020)Caesar, Bankiti, Lang, Vora, Liong, Xu, Krishnan, Pan, Baldan, and Beijbom]{caesar2020nuscenes}
Holger Caesar, Varun Bankiti, Alex~H Lang, Sourabh Vora, Venice~Erin Liong, Qiang Xu, Anush Krishnan, Yu Pan, Giancarlo Baldan, and Oscar Beijbom.
\newblock nuscenes: A multimodal dataset for autonomous driving.
\newblock In \emph{Proceedings of the IEEE/CVF conference on computer vision and pattern recognition}, pages 11621--11631, 2020.

\bibitem[Chen et~al.(2022)Chen, Wang, Darrell, and Ebrahimi]{chen2022contrastive}
Dian Chen, Dequan Wang, Trevor Darrell, and Sayna Ebrahimi.
\newblock Contrastive test-time adaptation.
\newblock In \emph{Proceedings of the IEEE/CVF Conference on Computer Vision and Pattern Recognition}, pages 295--305, 2022.

\bibitem[Chen et~al.(2019)Chen, Yang, Liang, and Urtasun]{chen2019learning}
Yun Chen, Bin Yang, Ming Liang, and Raquel Urtasun.
\newblock Learning joint 2d-3d representations for depth completion.
\newblock In \emph{Proceedings of the IEEE/CVF International Conference on Computer Vision}, pages 10023--10032, 2019.

\bibitem[Cheng et~al.(2020)Cheng, Wang, Guan, and Yang]{cheng2020cspn++}
Xinjing Cheng, Peng Wang, Chenye Guan, and Ruigang Yang.
\newblock Cspn++: Learning context and resource aware convolutional spatial propagation networks for depth completion.
\newblock In \emph{Proceedings of the AAAI Conference on Artificial Intelligence}, pages 10615--10622, 2020.

\bibitem[Choi et~al.(2022)Choi, Yang, Choi, and Yun]{choi2022improving}
Sungha Choi, Seunghan Yang, Seokeon Choi, and Sungrack Yun.
\newblock Improving test-time adaptation via shift-agnostic weight regularization and nearest source prototypes.
\newblock In \emph{ECCV}, pages 440--458. Springer, 2022.

\bibitem[Cicek and Soatto(2019)]{cicek2019unsupervised}
Safa Cicek and Stefano Soatto.
\newblock Unsupervised domain adaptation via regularized conditional alignment.
\newblock In \emph{Proceedings of the IEEE/CVF international conference on computer vision}, pages 1416--1425, 2019.

\bibitem[Dai et~al.(2017)Dai, Chang, Savva, Halber, Funkhouser, and Nie{\ss}ner]{dai2017scannet}
Angela Dai, Angel~X Chang, Manolis Savva, Maciej Halber, Thomas Funkhouser, and Matthias Nie{\ss}ner.
\newblock Scannet: Richly-annotated 3d reconstructions of indoor scenes.
\newblock In \emph{Proceedings of the IEEE conference on computer vision and pattern recognition}, pages 5828--5839, 2017.

\bibitem[Dou et~al.(2024)Dou, Yang, Liu, Loquercio, and Owens]{dou2024tactile}
Yiming Dou, Fengyu Yang, Yi Liu, Antonio Loquercio, and Andrew Owens.
\newblock Tactile-augmented radiance fields.
\newblock In \emph{Proceedings of the IEEE/CVF Conference on Computer Vision and Pattern Recognition}, 2024.

\bibitem[Eldesokey et~al.(2018)Eldesokey, Felsberg, and Khan]{eldesokey2018propagating}
Abdelrahman Eldesokey, Michael Felsberg, and Fahad~Shahbaz Khan.
\newblock Propagating confidences through cnns for sparse data regression.
\newblock \emph{arXiv preprint arXiv:1805.11913}, 2018.

\bibitem[Eldesokey et~al.(2020)Eldesokey, Felsberg, Holmquist, and Persson]{eldesokey2020uncertainty}
Abdelrahman Eldesokey, Michael Felsberg, Karl Holmquist, and Michael Persson.
\newblock Uncertainty-aware cnns for depth completion: Uncertainty from beginning to end.
\newblock In \emph{Proceedings of the IEEE/CVF Conference on Computer Vision and Pattern Recognition}, pages 12014--12023, 2020.

\bibitem[Fei et~al.(2019)Fei, Wong, and Soatto]{fei2019geo}
Xiaohan Fei, Alex Wong, and Stefano Soatto.
\newblock Geo-supervised visual depth prediction.
\newblock \emph{IEEE Robotics and Automation Letters}, 4\penalty0 (2):\penalty0 1661--1668, 2019.

\bibitem[Gaidon et~al.(2016)Gaidon, Wang, Cabon, and Vig]{gaidon2016virtual}
Adrien Gaidon, Qiao Wang, Yohann Cabon, and Eleonora Vig.
\newblock Virtual worlds as proxy for multi-object tracking analysis.
\newblock In \emph{Proceedings of the IEEE conference on computer vision and pattern recognition}, pages 4340--4349, 2016.

\bibitem[Ganin and Lempitsky(2015)]{ganin2015unsupervised}
Yaroslav Ganin and Victor Lempitsky.
\newblock Unsupervised domain adaptation by backpropagation.
\newblock In \emph{International conference on machine learning}, pages 1180--1189. PMLR, 2015.

\bibitem[Geiger et~al.(2013)Geiger, Lenz, Stiller, and Urtasun]{geiger2013vision}
Andreas Geiger, Philip Lenz, Christoph Stiller, and Raquel Urtasun.
\newblock Vision meets robotics: The kitti dataset.
\newblock \emph{The International Journal of Robotics Research}, 32:\penalty0 1231 -- 1237, 2013.

\bibitem[Grill et~al.(2020)Grill, Strub, Altch{\'e}, Tallec, Richemond, Buchatskaya, Doersch, Avila~Pires, Guo, Gheshlaghi~Azar, et~al.]{grill2020bootstrap}
Jean-Bastien Grill, Florian Strub, Florent Altch{\'e}, Corentin Tallec, Pierre Richemond, Elena Buchatskaya, Carl Doersch, Bernardo Avila~Pires, Zhaohan Guo, Mohammad Gheshlaghi~Azar, et~al.
\newblock Bootstrap your own latent-a new approach to self-supervised learning.
\newblock \emph{Advances in neural information processing systems}, 33:\penalty0 21271--21284, 2020.

\bibitem[Harris and Stephens(1988)]{harris1988combined}
Christopher~G. Harris and M.~J. Stephens.
\newblock A combined corner and edge detector.
\newblock In \emph{Alvey Vision Conference}, 1988.

\bibitem[Hu et~al.(2021)Hu, Wang, Li, Ning, Fan, and Gong]{hu2021penet}
Mu Hu, Shuling Wang, Bin Li, Shiyu Ning, Li Fan, and Xiaojin Gong.
\newblock Penet: Towards precise and efficient image guided depth completion.
\newblock In \emph{2021 IEEE International Conference on Robotics and Automation (ICRA)}, pages 13656--13662. IEEE, 2021.

\bibitem[Huang et~al.(2019)Huang, Fan, Cheng, Yi, Wang, and Li]{huang2019hms}
Zixuan Huang, Junming Fan, Shenggan Cheng, Shuai Yi, Xiaogang Wang, and Hongsheng Li.
\newblock Hms-net: Hierarchical multi-scale sparsity-invariant network for sparse depth completion.
\newblock \emph{IEEE Transactions on Image Processing}, 29:\penalty0 3429--3441, 2019.

\bibitem[Iwasawa and Matsuo(2021)]{iwasawa2021test}
Yusuke Iwasawa and Yutaka Matsuo.
\newblock Test-time classifier adjustment module for model-agnostic domain generalization.
\newblock In \emph{NIPS}, pages 2427--2440, 2021.

\bibitem[Jaritz et~al.(2018)Jaritz, De~Charette, Wirbel, Perrotton, and Nashashibi]{jaritz2018sparse}
Maximilian Jaritz, Raoul De~Charette, Emilie Wirbel, Xavier Perrotton, and Fawzi Nashashibi.
\newblock Sparse and dense data with cnns: Depth completion and semantic segmentation.
\newblock In \emph{2018 International Conference on 3D Vision (3DV)}, pages 52--60. IEEE, 2018.

\bibitem[Kam et~al.(2022)Kam, Kim, Kim, Park, and Lee]{kam2022costdcnet}
Jaewon Kam, Jungeon Kim, Soongjin Kim, Jaesik Park, and Seungyong Lee.
\newblock Costdcnet: Cost volume based depth completion for a single rgb-d image.
\newblock In \emph{European Conference on Computer Vision}, pages 257--274. Springer, 2022.

\bibitem[Kim et~al.(2021)Kim, Cho, Han, Panda, and Hong]{kim2021domain}
Youngeun Kim, Donghyeon Cho, Kyeongtak Han, Priyadarshini Panda, and Sungeun Hong.
\newblock Domain adaptation without source data.
\newblock \emph{IEEE Transactions on Artificial Intelligence}, 2\penalty0 (6):\penalty0 508--518, 2021.

\bibitem[Li et~al.(2020)Li, Yuan, Ling, Chi, Zhang, et~al.]{li2020multi}
Ang Li, Zejian Yuan, Yonggen Ling, Wanchao Chi, Chong Zhang, et~al.
\newblock A multi-scale guided cascade hourglass network for depth completion.
\newblock In \emph{Proceedings of the IEEE/CVF Winter Conference on Applications of Computer Vision}, pages 32--40, 2020.

\bibitem[Lin et~al.(2022)Lin, Cheng, Zhong, Zhou, and Yang]{lin2022dynamic}
Yuankai Lin, Tao Cheng, Qi Zhong, Wending Zhou, and Hua Yang.
\newblock Dynamic spatial propagation network for depth completion.
\newblock In \emph{Proceedings of the AAAI Conference on Artificial Intelligence}, pages 1638--1646, 2022.

\bibitem[Liu et~al.(2022)Liu, Agrawal, Chen, Hong, and Wong]{liu2022monitored}
Tian~Yu Liu, Parth Agrawal, Allison Chen, Byung-Woo Hong, and Alex Wong.
\newblock Monitored distillation for positive congruent depth completion.
\newblock In \emph{Computer Vision--ECCV 2022: 17th European Conference, Tel Aviv, Israel, October 23--27, 2022, Proceedings, Part II}, pages 35--53. Springer, 2022.

\bibitem[Liu et~al.(2021)Liu, Kothari, Van~Delft, Bellot-Gurlet, Mordan, and Alahi]{liu2021ttt++}
Yuejiang Liu, Parth Kothari, Bastien Van~Delft, Baptiste Bellot-Gurlet, Taylor Mordan, and Alexandre Alahi.
\newblock {TTT++}: When does self-supervised test-time training fail or thrive?
\newblock In \emph{NIPS}, pages 21808--21820, 2021.

\bibitem[Lopez-Rodriguez et~al.(2020)Lopez-Rodriguez, Busam, and Mikolajczyk]{lopez2020project}
Adrian Lopez-Rodriguez, Benjamin Busam, and Krystian Mikolajczyk.
\newblock Project to adapt: Domain adaptation for depth completion from noisy and sparse sensor data.
\newblock In \emph{Proceedings of the Asian Conference on Computer Vision}, 2020.

\bibitem[Ma et~al.(2019)Ma, Cavalheiro, and Karaman]{ma2019self}
Fangchang Ma, Guilherme~Venturelli Cavalheiro, and Sertac Karaman.
\newblock Self-supervised sparse-to-dense: Self-supervised depth completion from lidar and monocular camera.
\newblock In \emph{2019 International Conference on Robotics and Automation (ICRA)}, pages 3288--3295. IEEE, 2019.

\bibitem[McCormac et~al.(2016)McCormac, Handa, Leutenegger, and Davison]{mccormac2016scenenet}
John McCormac, Ankur Handa, Stefan Leutenegger, and Andrew~J Davison.
\newblock Scenenet rgb-d: 5m photorealistic images of synthetic indoor trajectories with ground truth.
\newblock \emph{arXiv preprint arXiv:1612.05079}, 2016.

\bibitem[Nathan~Silberman and Fergus(2012)]{Silberman:ECCV12}
Pushmeet~Kohli Nathan~Silberman, Derek~Hoiem and Rob Fergus.
\newblock Indoor segmentation and support inference from rgbd images.
\newblock In \emph{ECCV}, 2012.

\bibitem[Park et~al.(2020)Park, Joo, Hu, Liu, and So~Kweon]{park2020non}
Jinsun Park, Kyungdon Joo, Zhe Hu, Chi-Kuei Liu, and In So~Kweon.
\newblock Non-local spatial propagation network for depth completion.
\newblock In \emph{Computer Vision--ECCV 2020: 16th European Conference, Glasgow, UK, August 23--28, 2020, Proceedings, Part XIII 16}, pages 120--136. Springer, 2020.

\bibitem[Peng et~al.(2019)Peng, Bai, Xia, Huang, Saenko, and Wang]{peng2019moment}
Xingchao Peng, Qinxun Bai, Xide Xia, Zijun Huang, Kate Saenko, and Bo Wang.
\newblock Moment matching for multi-source domain adaptation.
\newblock In \emph{Proceedings of the IEEE/CVF international conference on computer vision}, pages 1406--1415, 2019.

\bibitem[Qiu et~al.(2019)Qiu, Cui, Zhang, Zhang, Liu, Zeng, and Pollefeys]{qiu2019deeplidar}
Jiaxiong Qiu, Zhaopeng Cui, Yinda Zhang, Xingdi Zhang, Shuaicheng Liu, Bing Zeng, and Marc Pollefeys.
\newblock Deeplidar: Deep surface normal guided depth prediction for outdoor scene from sparse lidar data and single color image.
\newblock In \emph{Proceedings of the IEEE/CVF Conference on Computer Vision and Pattern Recognition}, pages 3313--3322, 2019.

\bibitem[Qu et~al.(2020)Qu, Nguyen, and Taylor]{qu2020depth}
Chao Qu, Ty Nguyen, and Camillo Taylor.
\newblock Depth completion via deep basis fitting.
\newblock In \emph{Proceedings of the IEEE/CVF Winter Conference on Applications of Computer Vision}, pages 71--80, 2020.

\bibitem[Qu et~al.(2021)Qu, Liu, and Taylor]{qu2021bayesian}
Chao Qu, Wenxin Liu, and Camillo~J Taylor.
\newblock Bayesian deep basis fitting for depth completion with uncertainty.
\newblock In \emph{Proceedings of the IEEE/CVF international conference on computer vision}, pages 16147--16157, 2021.

\bibitem[Rho et~al.(2022)Rho, Ha, and Kim]{rho2022guideformer}
Kyeongha Rho, Jinsung Ha, and Youngjung Kim.
\newblock Guideformer: Transformers for image guided depth completion.
\newblock In \emph{Proceedings of the IEEE/CVF Conference on Computer Vision and Pattern Recognition}, pages 6250--6259, 2022.

\bibitem[Shin et~al.(2022)Shin, Tsai, Zhuang, Schulter, Liu, Garg, Kweon, and Yoon]{shin2022mm}
Inkyu Shin, Yi-Hsuan Tsai, Bingbing Zhuang, Samuel Schulter, Buyu Liu, Sparsh Garg, In~So Kweon, and Kuk-Jin Yoon.
\newblock Mm-tta: multi-modal test-time adaptation for 3d semantic segmentation.
\newblock In \emph{CVPR}, pages 16928--16937, 2022.

\bibitem[Shivakumar et~al.(2019)Shivakumar, Nguyen, Miller, Chen, Kumar, and Taylor]{shivakumar2019dfusenet}
Shreyas~S Shivakumar, Ty Nguyen, Ian~D Miller, Steven~W Chen, Vijay Kumar, and Camillo~J Taylor.
\newblock Dfusenet: Deep fusion of rgb and sparse depth information for image guided dense depth completion.
\newblock In \emph{2019 IEEE Intelligent Transportation Systems Conference (ITSC)}, pages 13--20. IEEE, 2019.

\bibitem[Silberman et~al.(2012)Silberman, Hoiem, Kohli, and Fergus]{silberman2012indoor}
Nathan Silberman, Derek Hoiem, Pushmeet Kohli, and Rob Fergus.
\newblock Indoor segmentation and support inference from rgbd images.
\newblock In \emph{European Conference on Computer Vision}, 2012.

\bibitem[Singh et~al.(2023)Singh, Ba, Sarker, Zhang, Kadambi, Soatto, Srivastava, and Wong]{singh2023depth}
Akash~Deep Singh, Yunhao Ba, Ankur Sarker, Howard Zhang, Achuta Kadambi, Stefano Soatto, Mani Srivastava, and Alex Wong.
\newblock Depth estimation from camera image and mmwave radar point cloud.
\newblock In \emph{Proceedings of the IEEE/CVF Conference on Computer Vision and Pattern Recognition}, pages 9275--9285, 2023.

\bibitem[Song et~al.(2023)Song, Lee, Kweon, and Choi]{song2023ecotta}
Junha Song, Jungsoo Lee, In~So Kweon, and Sungha Choi.
\newblock Ecotta: Memory-efficient continual test-time adaptation via self-distilled regularization.
\newblock In \emph{Proceedings of the IEEE/CVF Conference on Computer Vision and Pattern Recognition}, pages 11920--11929, 2023.

\bibitem[Sun et~al.(2020{\natexlab{a}})Sun, Kretzschmar, Dotiwalla, Chouard, Patnaik, Tsui, Guo, Zhou, Chai, Caine, et~al.]{sun2020scalability}
Pei Sun, Henrik Kretzschmar, Xerxes Dotiwalla, Aurelien Chouard, Vijaysai Patnaik, Paul Tsui, James Guo, Yin Zhou, Yuning Chai, Benjamin Caine, et~al.
\newblock Scalability in perception for autonomous driving: Waymo open dataset.
\newblock In \emph{Proceedings of the IEEE/CVF conference on computer vision and pattern recognition}, pages 2446--2454, 2020{\natexlab{a}}.

\bibitem[Sun et~al.(2019)Sun, Tzeng, Darrell, and Efros]{sun2019unsupervised}
Yu Sun, Eric Tzeng, Trevor Darrell, and Alexei~A Efros.
\newblock Unsupervised domain adaptation through self-supervision.
\newblock \emph{arXiv preprint arXiv:1909.11825}, 2019.

\bibitem[Sun et~al.(2020{\natexlab{b}})Sun, Wang, Liu, Miller, Efros, and Hardt]{sun2020test}
Yu Sun, Xiaolong Wang, Zhuang Liu, John Miller, Alexei Efros, and Moritz Hardt.
\newblock Test-time training with self-supervision for generalization under distribution shifts.
\newblock In \emph{ICML}, pages 9229--9248. PMLR, 2020{\natexlab{b}}.

\bibitem[Uhrig et~al.(2017)Uhrig, Schneider, Schneider, Franke, Brox, and Geiger]{uhrig2017sparsity}
Jonas Uhrig, Nick Schneider, Lukas Schneider, Uwe Franke, Thomas Brox, and Andreas Geiger.
\newblock Sparsity invariant cnns.
\newblock In \emph{2017 international conference on 3D Vision (3DV)}, pages 11--20. IEEE, 2017.

\bibitem[Upadhyay et~al.(2023)Upadhyay, Zhang, Ba, Yang, Gella, Jiang, Wong, and Kadambi]{upadhyay2023enhancing}
Rishi Upadhyay, Howard Zhang, Yunhao Ba, Ethan Yang, Blake Gella, Sicheng Jiang, Alex Wong, and Achuta Kadambi.
\newblock Enhancing diffusion models with 3d perspective geometry constraints.
\newblock \emph{ACM Transactions on Graphics (TOG)}, 42\penalty0 (6):\penalty0 1--15, 2023.

\bibitem[Van~der Maaten and Hinton(2008)]{van2008visualizing}
Laurens Van~der Maaten and Geoffrey Hinton.
\newblock Visualizing data using t-sne.
\newblock \emph{Journal of machine learning research}, 9\penalty0 (11), 2008.

\bibitem[Van~Gansbeke et~al.(2019)Van~Gansbeke, Neven, De~Brabandere, and Van~Gool]{van2019sparse}
Wouter Van~Gansbeke, Davy Neven, Bert De~Brabandere, and Luc Van~Gool.
\newblock Sparse and noisy lidar completion with rgb guidance and uncertainty.
\newblock In \emph{2019 16th international conference on machine vision applications (MVA)}, pages 1--6. IEEE, 2019.

\bibitem[Wang et~al.(2020)Wang, Shelhamer, Liu, Olshausen, and Darrell]{wang2020tent}
Dequan Wang, Evan Shelhamer, Shaoteng Liu, Bruno Olshausen, and Trevor Darrell.
\newblock Tent: Fully test-time adaptation by entropy minimization.
\newblock \emph{arXiv preprint arXiv:2006.10726}, 2020.

\bibitem[Wang et~al.(2022)Wang, Fink, Van~Gool, and Dai]{wang2022continual}
Qin Wang, Olga Fink, Luc Van~Gool, and Dengxin Dai.
\newblock Continual test-time domain adaptation.
\newblock In \emph{Proceedings of the IEEE/CVF Conference on Computer Vision and Pattern Recognition}, pages 7201--7211, 2022.

\bibitem[Wong and Soatto(2019)]{wong2019bilateral}
Alex Wong and Stefano Soatto.
\newblock Bilateral cyclic constraint and adaptive regularization for unsupervised monocular depth prediction.
\newblock In \emph{Proceedings of the IEEE/CVF Conference on Computer Vision and Pattern Recognition}, pages 5644--5653, 2019.

\bibitem[Wong and Soatto(2021)]{wong2021unsupervised}
Alex Wong and Stefano Soatto.
\newblock Unsupervised depth completion with calibrated backprojection layers.
\newblock In \emph{Proceedings of the IEEE/CVF International Conference on Computer Vision}, pages 12747--12756, 2021.

\bibitem[Wong et~al.(2020)Wong, Fei, Tsuei, and Soatto]{wong2020unsupervised}
Alex Wong, Xiaohan Fei, Stephanie Tsuei, and Stefano Soatto.
\newblock Unsupervised depth completion from visual inertial odometry.
\newblock \emph{IEEE Robotics and Automation Letters}, 5\penalty0 (2):\penalty0 1899--1906, 2020.

\bibitem[Wong et~al.(2021{\natexlab{a}})Wong, Cicek, and Soatto]{wong2021learning}
Alex Wong, Safa Cicek, and Stefano Soatto.
\newblock Learning topology from synthetic data for unsupervised depth completion.
\newblock \emph{IEEE Robotics and Automation Letters}, 6\penalty0 (2):\penalty0 1495--1502, 2021{\natexlab{a}}.

\bibitem[Wong et~al.(2021{\natexlab{b}})Wong, Fei, Hong, and Soatto]{wong2021adaptive}
Alex Wong, Xiaohan Fei, Byung-Woo Hong, and Stefano Soatto.
\newblock An adaptive framework for learning unsupervised depth completion.
\newblock \emph{IEEE Robotics and Automation Letters}, 6\penalty0 (2):\penalty0 3120--3127, 2021{\natexlab{b}}.

\bibitem[Wu et~al.(2023)Wu, Liu, Park, Soatto, Lao, and Wong]{wu2023augundo}
Yangchao Wu, Tian~Yu Liu, Hyoungseob Park, Stefano Soatto, Dong Lao, and Alex Wong.
\newblock Augundo: Scaling up augmentations for unsupervised depth completion.
\newblock \emph{arXiv preprint arXiv:2310.09739}, 2023.

\bibitem[Xu et~al.(2019)Xu, Zhu, Shi, Zhang, Bao, and Li]{xu2019depth}
Yan Xu, Xinge Zhu, Jianping Shi, Guofeng Zhang, Hujun Bao, and Hongsheng Li.
\newblock Depth completion from sparse lidar data with depth-normal constraints.
\newblock In \emph{Proceedings of the IEEE/CVF International Conference on Computer Vision}, pages 2811--2820, 2019.

\bibitem[Yang et~al.(2022)Yang, Ma, Zhang, Zhu, Yuan, and Owens]{yang2022touch}
Fengyu Yang, Chenyang Ma, Jiacheng Zhang, Jing Zhu, Wenzhen Yuan, and Andrew Owens.
\newblock Touch and go: Learning from human-collected vision and touch.
\newblock \emph{Neural Information Processing Systems (NeurIPS) - Datasets and Benchmarks Track}, 2022.

\bibitem[Yang et~al.(2023)Yang, Zhang, and Owens]{yang2023generating}
Fengyu Yang, Jiacheng Zhang, and Andrew Owens.
\newblock Generating visual scenes from touch.
\newblock \emph{International Conference on Computer Vision (ICCV)}, 2023.

\bibitem[Yang et~al.(2024)Yang, Feng, Chen, Park, Wang, Dou, Zeng, Chen, Gangopadhyay, Owens, and Wong]{yang2024unitouch}
Fengyu Yang, Chao Feng, Ziyang Chen, Hyoungseob Park, Daniel Wang, Yiming Dou, Ziyao Zeng, Xien Chen, Rit Gangopadhyay, Andrew Owens, and Alex Wong.
\newblock Binding touch to everything: Learning unified multimodal tactile representations.
\newblock In \emph{Proceedings of the IEEE/CVF Conference on Computer Vision and Pattern Recognition}, 2024.

\bibitem[Yang et~al.(2019)Yang, Wong, and Soatto]{yang2019dense}
Yanchao Yang, Alex Wong, and Stefano Soatto.
\newblock Dense depth posterior (ddp) from single image and sparse range.
\newblock In \emph{Proceedings of the IEEE/CVF Conference on Computer Vision and Pattern Recognition}, pages 3353--3362, 2019.

\bibitem[Youmin et~al.(2023)Youmin, Xianda, Matteo, Zheng, Guan, and Stefano]{youmin2023completionformer}
Zhang Youmin, Guo Xianda, Poggi Matteo, Zhu Zheng, Huang Guan, and Mattoccia Stefano.
\newblock Completionformer: Depth completion with convolutions and vision transformers.
\newblock \emph{arXiv preprint arXiv:2304.13030}, 2023.

\bibitem[Zhang and Funkhouser(2018)]{zhang2018deep}
Yinda Zhang and Thomas Funkhouser.
\newblock Deep depth completion of a single rgb-d image.
\newblock In \emph{Proceedings of the IEEE Conference on Computer Vision and Pattern Recognition}, pages 175--185, 2018.

\end{thebibliography}
}

\clearpage
\appendix
\begin{center} 
    {\LARGE{\textbf{Supplementary Materials} \\}}
\end{center}\section*{Summary of contents}
\begin{itemize}
    \item In Section A, we present the GPU time of each adaptation method to show the effectiveness of our method.
    \item In Section B, we present the preliminary observations with image and range inputs of varying sparsity.
    \item In Section C, we describe the datasets used.
    \item In Section D, we present the comparison of qualitative results on Waymo dataset's adverse weather sample from pretrained and adapted model to show the effectiveness of our adaptation under adverse weather conditions.
    \item In Section E, we present the hyperparameter settings for result reproduction and we elucidate evaluation details.
    \item In Section F, we provide a study on the learned proxy embedding with a visualization.
    \item In Section G, we present an ablation study of the loss components in our method.
    \item In Section H, we present the results on KITTI $\rightarrow$ VKITTI adaptation.
    \item In Section I, we present the results on a different source dataset (Waymo $\rightarrow$ VKITTI).
    \item In Section J, we show a qualitavive result of the preliminary observation.
\end{itemize}

\section*{A. Adaptation speed}
\label{sec:Speed}
We compare the GPU time of our adaptation method with the baselines (BN Adapt, CoTTA) on VKITTI in \tabref{tab:gpu_time}.

Compared to CoTTA, our adaptation method does not require multiple inferences to get the pseudo-prediction (derived from averaging teacher model predictions with different RGB augmentations) used to adapt the student model. Yet, our method requires an additional computation for the proxy embedding. Thus, the proxy layer's size relative to the model size causes the adaptation time difference.
For example, CoTTA reduced the total time by 38.9\% over ProxyTTA-fast on MSGCHN, which is a light-weight depth completion model. In this case, the proxy layer is relatively larger than in other models, where multiple inferences require less computation than the proxy layer. As a result, the total time is increased in MSGCHN.
However, for large models (NLSPN, CostDCNet), ProxyTTA reduced total time by 56.6\% over CoTTA; our proxy layer size is relatively smaller than the large models, while still improving performance by 26.52\%.
Compared to BN Adapt, our method requires additional parameters for the adaptation layer and the proxy layer. Hence, our method is 38.18\% slower in adaptation time, 19.36\% slower in evaluation time, and 33.16\% in total. Yet, our method improves errors by 15.67\% over BN Adapt.
 
\begin{table}[h!]
\centering
\setlength\tabcolsep{4pt}
\scriptsize
\begin{tabular}{lccccc}
\toprule
 Model & Method & Adaptation time & Evaluation time & Total time \\
\midrule
\multirow{2}{*}{MSGCHN} & CoTTA & 88.9 \textcolor{green}{(-38.9\%)} & 8.66 \textcolor{green}{(-1.0\%)} & 81.2 \textcolor{green}{(-41.3\%)} \\
& ProxyTTA-fast & 136.6 & 8.8 & 145.4 \\
\midrule
\multirow{3}{*}{NLSPN} & CoTTA & 717.5 \textcolor{red}{(+67.4\%)} & 75.3 \textcolor{green}{(-10.9\%)} & 792.8 \textcolor{red}{(+60.0\%)} \\
& BN Adapt & 185.0 \textcolor{green}{(-20.8\%)} & 82.8 \textcolor{green}{(-0.8\%)} & 267.8 \textcolor{green}{(-15.6\%)} \\ 

& ProxyTTA-fast & 168.2 \textcolor{green}{(-28.0\%)} &83.4 \textcolor{green}{(-0.1\%)} &251.6 \textcolor{green}{(-20.66\%)} \\
& ProxyTTA &  233.6  & 83.5 & 317.1  \\ 
\midrule
\multirow{3}{*}{CostDCNet} & CoTTA & 329.1 \textcolor{red}{(+78.2\%)} & 33.6 \textcolor{green}{(-51.0\%)} & 369.1 \textcolor{red}{(+43.2\%)}\\
& BN Adapt & 82.1 \textcolor{green}{(-55.5\%)} & 42.5 \textcolor{green}{(-37.9\%)} & 125.6 \textcolor{green}{(-50.8\%)} \\ 
& ProxyTTA-fast & 141.9 \textcolor{green}{(-23.2\%)} &68.7 \textcolor{red}{(+0.3\%)} &210.6 \textcolor{green}{(-16.8\%)} \\
& ProxyTTA & 184.7 & 68.5 & 253.2 \\ 
\bottomrule
\end{tabular}
\caption{\textit{GPU time for various methods and models, tested on Virtual KITTI}. Time is in milliseconds (ms). `Adaptation time' denotes the time required to adapt (or train) each method for a single test data point. `Evaluation time' denotes the time taken to test each method for a test data instance.`Total time' is the sum of the Adaptation and Evaluation times.}
\label{tab:gpu_time}
\end{table}
\section*{B. Further observations on image/range inputs}
\label{sec:add_preliminary}
We present additional preliminary observations of the image and range sensor inputs with varying sparsity.
Since previous works~\cite{park2020non, cheng2020cspn++} state that the depth completion model propagates the sparse depth to the dense depth guided by image features, one can raise a question on our preliminary results in the main paper without the lidar input, such as there's no sparse point to propagate to the near pixels.
We clarify that the results are intended to highlight the domain distrepancy.
Therefore, we show additional results with 1\%, 5\%, and 10\% of sparse points in the range input on indoor datasets, as shown in Table \ref{table:exp:extra_preliminary}.
As we increase the range points, the performance is improved yet still worse than the sparse-depth-only results in Tab. \ref{table:exp:preliminary}.

\begin{table*}[t]
% \addtolength{\tabcolsep}{0.5pt}
\scriptsize
\centering
\setlength\tabcolsep{2pt}
\vspace{-2mm}
\resizebox{0.95\textwidth}{!}{%
\begin{tabular}{lcccccccccccc}
\toprule
Method               & \multicolumn{2}{c}{MSG-CHN}         & \multicolumn{2}{c}{NLSPN}            & \multicolumn{2}{c}{CostDCNet}       & \multicolumn{2}{c}{MSG-CHN}           & \multicolumn{2}{c}{NLSPN}             & \multicolumn{2}{c}{CostDCNet}       \\
\midrule
Dataset              & \multicolumn{6}{c}{{VOID$\rightarrow$NYUv2}}                                                              & 
\multicolumn{6}{c}{{VOID$\rightarrow$ScanNet}}                                                               \\
\cmidrule{2-7}  
\cmidrule{8-13}
                     & MAE              & RMSE             & MAE              & RMSE              & MAE              & RMSE             & MAE               & RMSE              & MAE               & RMSE              & MAE              & RMSE             \\
\midrule
Image + sparse depth (1\%)             &   1643.34 & 2177.71 &    602.17 &  858.19 & 809.36 & 1144.91                 & 1597.41 & 2240.43   &490.13 &  738.77 &    665.57 &  982.32 \\
\midrule
Image + sparse depth (5\%)      &  996.54 & 1599.14  &379.45  & 638.55  & 427.69  & 736.23 & 809.38 & 1455.69 &  240.55   & 441.70 & 337.39 &  620.53  \\
\midrule
Image + sparse depth (10\%)  & 785.65 & 1376.93 &327.41  & 591.99 &   339.31 &  622.75    &581.93 & 1165.63&191.75 &  379.10 & 264.66  & 516.74        \\
\midrule
Sparse depth only    & \textbf{734.13}  & \textbf{1046.28} & \textbf{237.47}  & \textbf{402.47}   & \textbf{147.76} & \textbf{354.57} & \textbf{211.86}  & \textbf{444.62}  & \textbf{162.29}  & \textbf{276.29}  & \textbf{88.25}  & \textbf{205.46} \\
\bottomrule
\end{tabular}
}
\caption{\textit{Model sensitivity to input modalities with varying sparsity.}  % Using just sparse depth improves over both modalities in test domain.
}
\label{table:exp:extra_preliminary}
\end{table*}

\section*{C. Datasets}
\label{sec:datasets}

\textbf{KITTI} \cite{geiger2013vision} is composed of calibrated RGB images with synchronized point clouds from Velodyne lidar, inertial, and GPS information, and from more than 61 driving scenes. There are $\approx$80K raw image frames and associated sparse depth maps, both with $\approx$5\% density, available for depth completion \cite{uhrig2017sparsity}. Semi-dense depth is available for the lower 30\% of the image space, and 11 neighboring raw lidar scans comprise the ground-truth depth. We did not use a test or validation set, and the training set contains $\approx$86K single images.

\textbf{VOID} \cite{wong2020unsupervised} contains synchronized 640$\times$480 RGB images and sparse depth maps from indoor scenes of laboratories and classrooms and from outdoor scenes of gardens. Sparse depth maps (of $\approx$0.5\% density and containing $\approx$1,500 sparse dense points) are obtained by the VIO system XIVO \cite{fei2019geo}, and dense ground-truth depth maps are obtained by active stereo. VOID uses rolling shutter to capture challenging 6 DoF motion for 56 sequences - as opposed to KITTI's typically planar motion. We use a training set of $\approx$46K images to prepare the model.

\textbf{NYUv2} \cite{silberman2012indoor} contains 372K synchronized 640$\times$480 RGB images and depth maps (via Microsoft Kinect) from 464 indoor scenes of household, office, and commercial types. To generate sparse depth maps in the style of SLAM/VIO, we used the Harris corner detector \cite{harris1988combined} to sample $\approx$1,500 points from the depth maps. 
We use a set of 654 test set images for adaptation.

\textbf{ScanNet}~\cite{dai2017scannet} contains 2.5 million images and dense depth maps for 1,513 indoor scenes. To generate sparse depth maps in the style of SLAM/VIO, we used the Harris corner detector~\cite{harris1988combined} to sample $\approx$1,500 points from the depth maps. We use a set of $\approx$21K test images for adaptation.

\textbf{Virtual KITTI (VKITTI)}~\cite{gaidon2016virtual} contains $\approx$17K   1242$\times$375 images from 35 synthetic videos created by applying 7 variations in weather, lighting, or camera angle to each of 5 cloned KITTI \cite{uhrig2017sparsity} videos. There exists a large domain gap between RGB images from VKITTI and KITTI, even though the virtual worlds created in Unity by \cite{gaidon2016virtual} are similar to KITTI scenes. Thus, we only use the dense depth maps of VKITTI to avoid the domain gap in photometric varations. The sparse depth maps are obtained by simulating KITTI's lidar-generated sparse depth measurements such that the marginal distribution of VKITTI's sparse points mimics that of KITTI's. We use a set of $\approx$2,300 test images for the adaptation.

\begin{table}[h!]
             \centering 
             \footnotesize
             \renewcommand{\arraystretch}{1.2}
             \begin{tabular}{cccccc} \hline 
                  Dataset &  Learning Rate &  $w_{sm}$&  $w_{z}$&  $w_{proxy}$ & Inner Iter.\\\hline\hline
 \multicolumn{6}{c}{MSG-CHN}\\\hline
                  VKITTI&  2e-3&  1.0&  1.0&  0.2& 1\\ 
                  VKITTI-FOG&  5e-3&  3.0&  1.0&  0.1& 1\\  
        nuScenes&  3e-3&  9.0&  1.0&  0.2& 1\\ 
                  SceneNet&  2e-3&  8.0&  1.0&  0.1& 3\\
                  NYUv2&  2e-4&  0.8&  1.0&  0.4& 3\\ 
                  ScanNet&  5e-3&  8.0&  1.0&  0.3& 3\\\hline
 \multicolumn{6}{c}{NLSPN}\\\hline
 VKITTI& 2e-3& 0.8& 1.0& 0.4&1\\
 VKITTI-FOG& 1e-3& 1.0& 1.0& 0.2&1\\
 nuScenes& 1e-3& 1.0& 1.0& 0.1&1\\
 SceneNet& 2e-3& 0.7& 1.0& 2.0&3\\
 NYUv2& 4e-3& 5.0& 1.0& 1.0&3\\
 ScanNet& 1e-4& 2.0& 1.0& 0.3&3\\ \hline 
                  \multicolumn{6}{c}{CostDCNet}\\\hline
 VKITTI& 4e-3& 4.5& 1.0& 0.1&1\\
 VKITTI-FOG& 5e-3& 3.0& 1.0& 0.04&1\\  
                  nuScenes&  5e-3&  3.0&  1.0&  0.1& 1\\ 
                  SceneNet&  7e-3&  2.0&  1.0&  0.2& 3\\ 
                  NYUv2&  6e-3&  4.0&  1.0&  0.1& 3\\ 
 ScanNet& 3e-3& 1.0& 1.0& 0.2&3\\\hline
             \end{tabular}
             \caption{\textit{Hyperparameters.} For MSG-CHN, NLSPN, and CostDCNet methods for initialization, preparation, and adaptation.}
             \label{tab:hyperparameter}
\end{table}

\textbf{nuScenes}~\cite{caesar2020nuscenes} consists of 1600$\times$900 calibrated RGB images and synchronized sparse point clouds, 27.4K images from 1000 outdoor driving scenes for training, and 5.8K images from 150 scenes for testing. We set up the ground truth for the test images by merging projected sparse depth from forward-backward frames. The setup code will be released to clarify further details and reproducibility.

\begin{figure*}[t]
    \centering
    \includegraphics[width=0.86\linewidth]{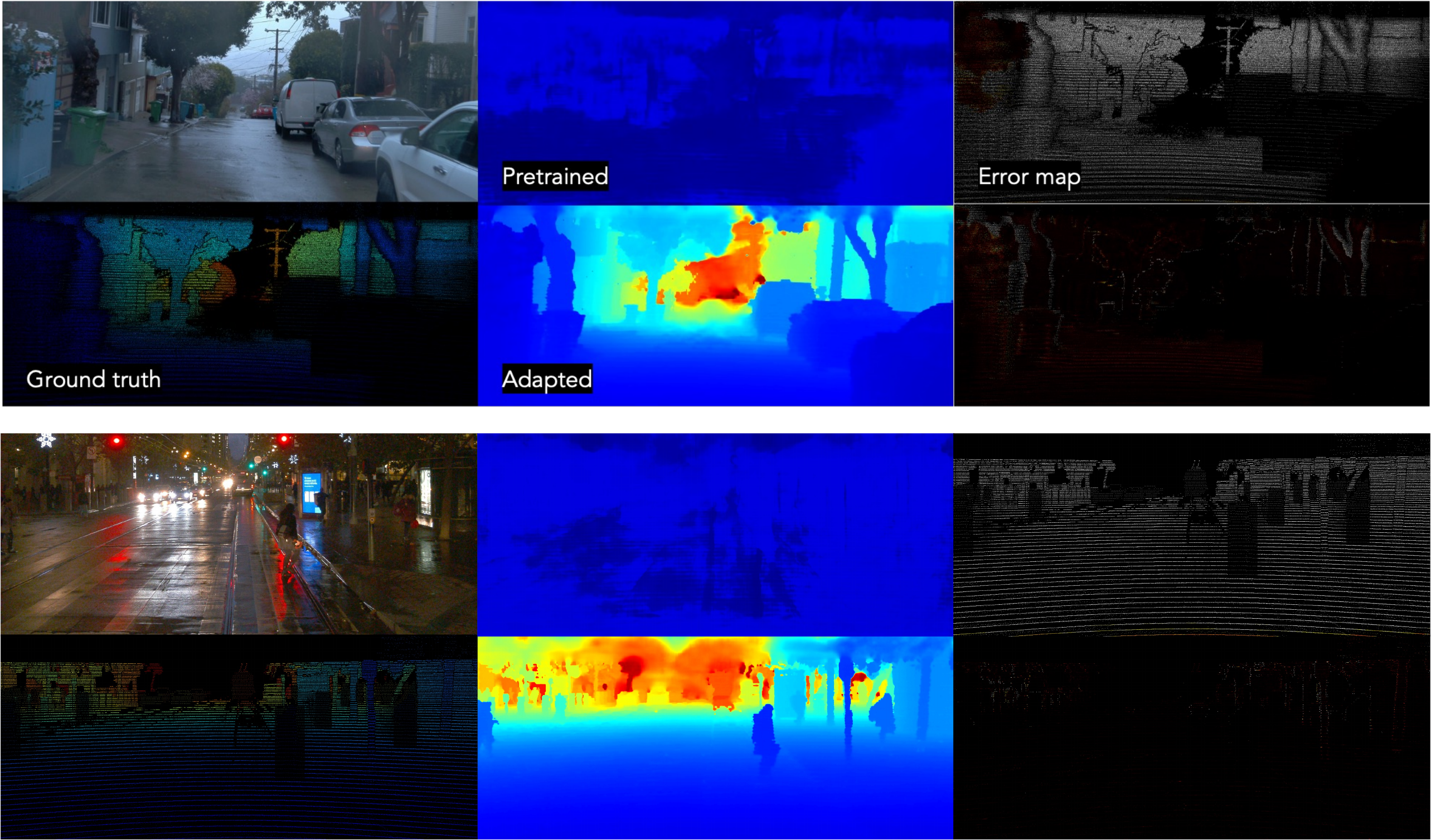}
    \caption{\textit{Qualitative results on Waymo. For outdoor adaptation scenarios, ProxyTTA can adapt under the adverse weather condition, such as raining condition (top row) and low-illumination (bottom row).}
    }
    \vspace{-3.5mm}
    \label{fig:adversarial}
\end{figure*}

\textbf{SceneNet} \cite{mccormac2016scenenet} contains 5 million 320$\times$240 RGB images and depth maps from indoor trajectories of randomly arranged rooms. We use a single split (out of 17 available) containing 1000 subsequences of 300 images each, generated by recording the same scene over a trajectory. Because there are no sparse depth maps provided, we sampled  from the depth map via Harris corner detector \cite{harris1988combined} to mimic the sparse depth produced by SLAM/VIO. The final 375 corners are obtained by using k-means to subsample the resulting points, representing 0.49\% of the total pixels. We use a set of $\approx$2,300 test images for adaptation.

\textbf{Waymo Open Dataset}~\cite{sun2020scalability} contains 1920$\times$1280 RGB images and lidar scans from autonomous vehicles. The training set contains $\approx$158K images from 798 scenes and the validation set $\approx$40K images from 202 scenes, collected at 10Hz. Objects are annotated across the full 360$^{\circ}$ field. We obtain our validation set by sampling from the whole validation dataset every 0.6 seconds.
Range sensor inputs are obtained by projecting the top lidar's point cloud scan to the camera frame. We obtained the ground truth by projecting 10 forward and backward frames from front lidar and top lidar to the image frame, which approximately counts for 1 second of capture.
To assume that the reprojected scenes are static, we removed the moving objects in the scenes using object annotations. Also, outlier removal is utilized for filtering out errorenous depth points.

\begin{table*}[t!]
\scriptsize
\centering
\setlength\tabcolsep{3pt}
\resizebox{0.95\textwidth}{!}{
\begin{tabular}{cl cccc cccc cccc}
    \toprule
    &  & & &  \multicolumn{2}{c}{KITTI $\rightarrow$ Waymo} & \multicolumn{2}{c}{KITTI $\rightarrow$ VKITTI-FOG} & \multicolumn{2}{c}{KITTI $\rightarrow$ nuScenes}\\
    \midrule
    Method & $\ell_z$ & $\ell_{sm}$& $\ell_\text{proxy}$ & MAE & RMSE & MAE & RMSE & MAE & RMSE \\
    \midrule 
    \midrule 
    \multirow{3}{*}{MSG-CHN} & \ding{51}& & & 951.25$\pm$3.14 & 3512.07$\pm$6.40 & 978.84$\pm$3.36 & 3561.40$\pm$15.48 & 3164.46$\pm$11.32&6453.54$\pm$17.31 \\
    &\ding{51} &\ding{51} & & \textit{613.01$\pm$1.99}&\textit{1935.43$\pm$9.14}& \textit{732.61$\pm$6.02} & \textit{3113.11$\pm$21.78} & \textit{2865.15$\pm$9.96} &\textit{6144.48$\pm$24.14} \\
    & \ding{51}&\ding{51} &\ding{51} & \textbf{608.91$\pm$1.74}&\textbf{1921.83$\pm$2.54} & \textbf{728.24}$\pm$\textbf{3.73} & \textbf{3087.36}$\pm$\textbf{15.92} & \textbf{2834.08$\pm$17.64}&\textbf{6096.56$\pm$21.08} \\ 
    \midrule
    \multirow{3}{*}{NLSPN} &\ding{51} & & & 837.66$\pm$ 8.73& 3668.94$\pm$ 25.90& 715.86$\pm$26.36 & \textit{3034.21$\pm$ 57.65} &  5076.83$\pm$53.85 & 9710.88$\pm$ 89.76\\
     & \ding{51}&\ding{51} & & \textit{489.46$\pm$5.45}&\textit{1613.66$\pm$30.04} & \textit{705.14$\pm$16.86} & 3059.64$\pm$97.85 & \textit{2783.61$\pm$159.62}&\textit{6313.4$\pm$276.09}\\
     &\ding{51} &\ding{51} &\ding{51} & \textbf{477.28$\pm$3.32}&\textbf{1598.64$\pm$18.95} & \textbf{686.91}$\pm$\textbf{22.14}  & \textbf{2666.70$\pm$56.64} 
     & \textbf{2589.25}$\pm$\textbf{59.03}&\textbf{6006.18}$\pm$\textbf{90.66}\\
     \midrule  
     \multirow{3}{*}{CostDCNet} & \ding{51}& & & 816.33$\pm$32.01 & 3431.96$\pm$55.34 & 807.62$\pm$69.12 & 3254.83$\pm$179.90 & 3135.11$\pm$81.76&7596.49$\pm$159.16 \\ 
     &\ding{51} &\ding{51} & & \textit{469.52$\pm$2.54}&\textit{1594.38$\pm$6.10} & \textit{516.93$\pm$1.62}&\textit{2751.21$\pm$17.42} & \textit{2067.42$\pm$10.23}&\textbf{5487.85$\pm$37.21} \\
     &\ding{51} &\ding{51} &\ding{51} & \textbf{466.44$\pm$1.63}&\textbf{1580.38$\pm$11.48}& \textbf{512.72}$\pm$\textbf{0.74}&\textbf{2735.01}$\pm$\textbf{3.53} & \textbf{2062.28$\pm$11.24} &\textit{5509.96$\pm$23.41} \\
    \midrule
    &  & & &  \multicolumn{2}{c}{VOID $\rightarrow$ NYUv2} & \multicolumn{2}{c}{VOID $\rightarrow$ SceneNet} & \multicolumn{2}{c}{VOID $\rightarrow$ ScanNet} \\ 

    \midrule 
    \multirow{3}{*}{MSG-CHN}  & \ding{51}& & & \textit{971.64$\pm$66.86} & \textit{1291.45$\pm$45.67} & 242.11$\pm$4.24& 491.48$\pm$10.49 & 462.95$\pm$34.84&659.9$\pm$37.93 \\
    &\ding{51} &\ding{51} & &  1005.49$\pm$25.97  & 1329.76$\pm$25.01 & \textit{194.60$\pm$3.64} & \textit{425.16$\pm$10.58} & \textit{330.20$\pm$48.46} & \textit{503.73$\pm$57.14} \\
    & \ding{51}&\ding{51} &\ding{51} & \textbf{699.60}$\pm$\textbf{6.00} & \textbf{1120.37}$\pm$\textbf{9.76} & \textbf{192.74}$\pm$\textbf{1.72} & \textbf{424.49}$\pm$\textbf{4.58} & \textbf{302.21$\pm$4.10} & \textbf{480.08$\pm$8.03} \\ 
    \midrule
    \multirow{3}{*}{NLSPN} & \ding{51}& & & 145.72 $\pm$6.55 & 271.78$\pm$ 9.91& 130.49$\pm$13.64 & \textit{337.14$\pm$28.38} & 112.38$\pm$1.72 & 234.60$\pm$3.46 \\
     &\ding{51} &\ding{51} & & \textit{128.17$\pm$4.13} & \textit{240.97$\pm$3.86} & \textit{118.65$\pm$2.24}&337.63$\pm$2.58 & \textit{77.84$\pm$0.28} & \textit{169.81$\pm$0.50} \\ 
     &\ding{51} &\ding{51} &\ding{51} & \textbf{124.41}$\pm$\textbf{2.27} & \textbf{240.73}$\pm$\textbf{5.72} &\textbf{113.93$\pm$1.49}&\textbf{333.41$\pm$4.32}& \textbf{74.77$\pm$0.31} & \textbf{166.61$\pm$0.45} \\
     \midrule  
     \multirow{3}{*}{CostDCNet} & \ding{51}& & &   152.43$\pm$13.07&432.20$\pm$54.51&213.4$\pm$19.52&597.22$\pm$49.78	 &91.13$\pm$1.40&286.17$\pm$9.07   \\
     &\ding{51} &\ding{51} & & \textit{101.31$\pm$1.67} & \textit{217.77$\pm$6.00} &\textit{134.51$\pm$4.23}&\textit{360.33$\pm$9.67} & \textit{69.02$\pm$0.51}&\textit{164.90$\pm$2.38}  \\
     &\ding{51} &\ding{51} &\ding{51} &   \textbf{95.87$\pm$2.16}&\textbf{203.83$\pm$4.72} &\textbf{125.75$\pm$1.93}&\textbf{357.12$\pm$4.13} &\textbf{68.17$\pm$0.44} &\textbf{162.35$\pm$1.12}   \\
    \midrule
\end{tabular}
}
\caption{\textit{Ablation study of each loss term.} Note that NLSPN and CostDCNet update the adaptation layer and batch normalization layers, yet MSGCHN only updates the adaptation layer.}
\vspace{-3mm}
\label{tab:experiments:ablation}
\end{table*}

\section*{D. Qualitative results on adverse weather conditions}

Typically, in real-world scenarios, most systems will encounter non-ideal sensing conditions, which will degrade performance. For example, existing pretrained depth (completion) models will fail under adverse weather conditions, such as nighttime (low-illumination) or rain. To address such failure modes of existing models, we adapt CostDCNet using Proxy-TTA. We demonstrate this capability in Fig. \ref{fig:adversarial}, where we improve over the pretrained model significantly as shown in the range and error map visualizations.

\section*{E. Implementation details}

\textbf{Hyperparameter.} We specifically note the hyperparameters of three methods for initialization, preparation, and adaptation on Table \ref{tab:hyperparameter}.

\noindent\textbf{Epochs and training details} Adaptation occurs in a single epoch, with `the number of iterations per data point' (\textit{inner-iter}) specified in Tab.~\ref{tab:hyperparameter}. During initialization and preparation stages, the adaptation and proxy layers are trained for 6 epochs.
Batch sizes for all methods are: 48 for preparation stage, 16 for initialization and adaptation stages, with the exception of ScanNet~[6], using a batch size of 36.
To prevent collapse during preparation stage, we follow the protocol of \cite{grill2020bootstrap}; we exploit the projection / prediction layers and divide online / target branch, and update target projection layer with exponential moving average of online branch. We used embedding dimension and hidden dimension of 512 for MSGCHN, and 1024 for CostDCNet and NLSPN. The learning rates for initialization and preparation stage will be released with the code release.

\noindent\textbf{Evaluation.}
We evaluate our adaptation models on bottom-cropped regions in the outdoor dataset, where the sparse depth exists.
For outdoor dataset, models are evaluated on the bottom cropped region of the test split, $1242\times240$ for Virtual KITTI, and $1600\times544$ for nuScenes. For indoor dataset, we evaluated the models on the entire region. The definition of the error metrics in  evaluation are described in Table \ref{tab:error_metrics}. We evaluate our model on depth range from 0.0 to 80.0 meters for the ourdoor, and 0.2 to 5.0 meters for the indooor.

\begin{table}[h]
\centering
\footnotesize
\setlength\tabcolsep{18pt}
\begin{tabular}{l l}
    \midrule
        Metric & Definition \\ \midrule
        MAE &$\frac{1}{|\Omega|} \sum_{x\in\Omega} |\hat d(x) - d_{gt}(x)|$ \\
        RMSE & $\big(\frac{1}{|\Omega|}\sum_{x\in\Omega}|\hat d(x) - d_{gt}(x)|^2 \big)^{1/2}$ \\
        % iMAE & $\frac{1}{|\Omega|} \sum_{x\in\Omega} |1/ \hat d(x) - 1/d_{gt}(x)|$ \\
        % iRMSE& $\big(\frac{1}{|\Omega|}\sum_{x\in\Omega}|1 / \hat d(x) - 1/d_{gt}(x)|^2\big)^{1/2}$ \\ 
        \midrule
    \end{tabular}
    \caption{
        \textit{Error metrics.} $d_{gt}$ denotes the ground-truth depth.
    }
    \vspace{-1.6em}
\label{tab:error_metrics}
\end{table}

\section*{F. Discussion on learned proxy embeddings}

Here, we provide the t-SNE visualization of image \& sparse depth and proxy embedding from source and target.

\begin{figure}[h]
    \centering
    \includegraphics[width=\linewidth]{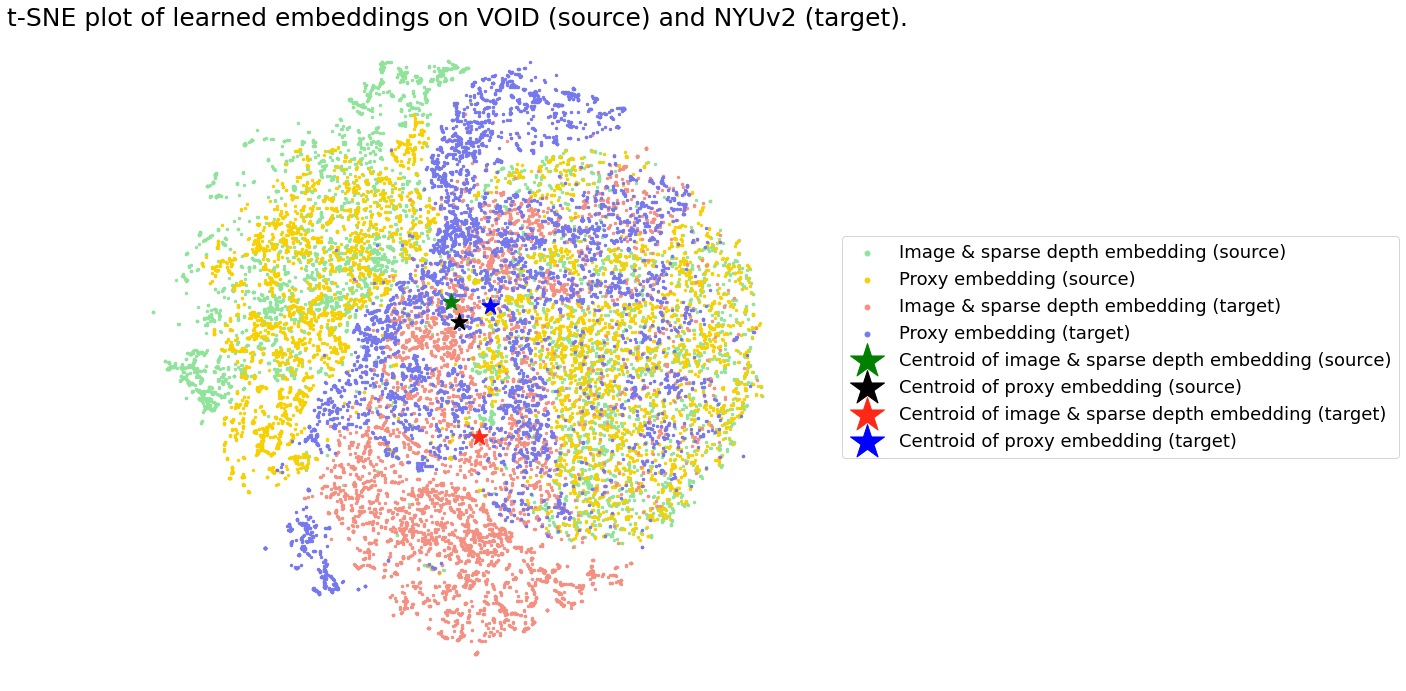}
    \caption{\textit{t-SNE plot of learned embeddings on VOID and NYUv2.}
    }
    \vspace{-3.5mm}
    \label{fig:t-sne}
\end{figure}

Fig. \ref{fig:t-sne} shows the embeddings visualized by t-SNE, where the target domain proxy embeddings' centroid is closer to that of source's proxy and image \& sparse depth embeddings, than to the centroid of target's image \& sparse depth embeddings, highlighting effectiveness of proxy embedding for adaptation. 

\begin{table}[t]
\scriptsize
\centering
\setlength\tabcolsep{3pt}
\resizebox{0.5\textwidth}{!}{%
\begin{tabular}{cl cccc}
    \toprule
    & \multicolumn{3}{c}{KITTI $\rightarrow$ VKITTI}\\
    \midrule
    Method & {} & MAE & RMSE \\
    \midrule 
    \midrule 
    \multirow{4}{*}{MSG-CHN} & Pretrained & 2433.46 & 6675.16 \\
    & CoTTA & \textit{839.19$\pm$12.78} & \textit{3625.38$\pm$39.35} \\ 
    & ProxyTTA-fast (Ours)& \textbf{800.88}$\pm$\textbf{1.86} & \textbf{3268.26}$\pm$\textbf{4.12} \\ 
    \midrule
    \multirow{6}{*}{NLSPN} &  Pretrained & 1469.19  &  8060.97  \\ 
     & BN Adapt & 1016.87$\pm$8.84 & \textit{3453.00$\pm$3.21} \\
     & BN Adapt, $\ell_z$, $\ell_{sm}$ & 855.12$\pm$14.56 & 3516.85$\pm$58.63 \\
     & CoTTA & \textit{775.09$\pm$3.63}&3585.37$\pm$13.31 \\  
     & ProxyTTA-fast & 849.43$\pm$3.61 & 3540.44$\pm$3.57 \\
     & ProxyTTA (Ours)& \textbf{639.19$\pm$5.68}&\textbf{2934.36$\pm$33.80} \\
     \midrule  
     \multirow{7}{*}{CostDCNet} &  Pretrained &   {845.35}  & {3774.01} \\ 
     & BN Adapt & 1248.35$\pm$0.25 & 4267.64$\pm$0.62 \\
     & BN Adapt, $\ell_z$, $\ell_{sm}$ & 1016.87$\pm$8.84 & 3453.00$\pm$3.21 \\
     & CoTTA & \textit{698.42$\pm$9.93}&\textit{3324.59$\pm$30.21} \\   
     & ProxyTTA-fast & 822.49$\pm$13.55 & 3331.24$\pm$55.30 \\  
     & ProxyTTA (Ours)& \textbf{639.91$\pm$8.92}&\textbf{2951.21$\pm$30.93} \\
    \toprule
\end{tabular}
    }
\vspace{-3mm}

\caption{\textit{Additional results for test-time adaptation for depth completion on KITTI $\rightarrow$ VKITTI.}% \textbf{Bold} denotes best and \textit{italics} runner-up.
}
\vspace{-7mm}
\label{tab:experiments:ablation:vkitti}
\end{table}

\begin{table*}[h!]
\scriptsize
\centering
\setlength\tabcolsep{2pt}
\vspace{-2mm}
\resizebox{0.95\textwidth}{!}{%
\begin{tabular}{lcccccccccccc}
\toprule
Method               & \multicolumn{2}{c}{MSG-CHN}         & \multicolumn{2}{c}{NLSPN}            & \multicolumn{2}{c}{CostDCNet}       & \multicolumn{2}{c}{MSG-CHN}           & \multicolumn{2}{c}{NLSPN}             & \multicolumn{2}{c}{CostDCNet}       \\
\midrule
Dataset              & \multicolumn{6}{c}{{VOID$\rightarrow$NYUv2}}                                                              & 
\multicolumn{6}{c}{{VOID$\rightarrow$ScanNet}}                                                               \\
\cmidrule{2-7}  
\cmidrule{8-13}
                     & MAE              & RMSE             & MAE              & RMSE              & MAE              & RMSE             & MAE               & RMSE              & MAE               & RMSE              & MAE              & RMSE             \\
\midrule
Image only           & 2072.78          & 2462.63          & 969.14           & 1228.44           &   1359.16 & 1619.40          & 2001.90           & 2451.681          & 899.41           & 1151.12          & 1216.17         & 1459.46         \\
\midrule
Sparse depth only    & \textbf{734.13}  & \textbf{1046.28} & \textbf{237.47}  & \textbf{402.47}   & \textbf{147.76} & \textbf{354.57} & \textbf{211.86}  & \textbf{444.62}  & \textbf{162.29}  & \textbf{276.29}  & \textbf{88.25}  & \textbf{205.46} \\
\midrule
Image + sparse depth & 1040.93          & 1528.98          & 387.36           & 704.66            & 189.10           & 446.71           & 316.646           & 698.633           & 232.332           & 431.199           & 144.311          & 458.692          \\
\midrule
Dataset              & \multicolumn{6}{c}{KITTI$\rightarrow$Waymo}                                                                     & \multicolumn{6}{c}{KITTI$\rightarrow$nuScenes}                                                                      \\
\midrule
Image only           &12766.791&18324.83&18829.96&24495.73&13598.50& 18376.15       & 11823.061         & 17244.44          & 15835.04          & 22613.78          & 12794.65        & 16744.15        \\
\midrule
Sparse depth only    &\textbf{861.13}&\textbf{2706.75}&1290.28&3571.26&1210.93& 3102.49         &  3943.97         & 7306.33        &  \textbf{2540.58} & 6203.66 &  \textbf{2996.28} & 6773.06    \\
\midrule
Image + sparse depth &1103.33&2969.39&\textbf{1173.26}&\textbf{3092.02}&\textbf{1084.18}&\textbf{2819.42} &\textbf{ 3331.82} &   \textbf{6449.09} & 2656.61           & \textbf{6146.59}         &  3064.72          & \textbf{6630.65} \\ 
\bottomrule
\end{tabular}
}
\caption{\textit{Model sensitivity to input modalities.}  % Using just sparse depth improves over both modalities in test domain.
Depth completion networks have a high reliance on sparse depth modality. Performing inference in a novel domain without the RGB image, i.e., using just sparse depth as input, can improve over using both data modalities.
}
\vspace{-3mm}
\label{table:exp:preliminary}
\end{table*}

\begin{table}[h]
\scriptsize
\centering
\setlength\tabcolsep{3pt}
\resizebox{0.5\textwidth}{!}{%
\begin{tabular}{cl cccc}
    \toprule
    &  & \multicolumn{2}{c}{Waymo $\rightarrow$ VKITTI-FOG}\\
    \midrule
    Method & {} & MAE & RMSE \\
    \midrule 
    \midrule 
    \multirow{3}{*}{MSG-CHN} & Pretrained & 1473.14 & 4676.19 \\
    & CoTTA &\textit{1348.02$\pm$38.03}&\textit{4016.67$\pm$28.16} \\ 
    & ProxyTTA-fast (Ours)& \textbf{1052.78$\pm$5.74}&\textbf{3891.05$\pm$17.34} \\ 
    \midrule
    \multirow{4}{*}{NLSPN} &  Pretrained & 2734.27  &  \textit{37621.10}  \\
     & BN Adapt, $\ell_z$, $\ell_{sm}$ & \textit{1205.96$\pm$40.14}&3857.88$\pm$101.15 \\
     & CoTTA & 2485.66$\pm$18.05&6307.96$\pm$48.64\\  
     & ProxyTTA (Ours)& \textbf{808.16$\pm$7.86}&\textbf{3536.58$\pm$91.15} \\
     \midrule  
     \multirow{4}{*}{CostDCNet} &  Pretrained &   {1261.00}  & {4360.37} \\ 
     & BN Adapt, $\ell_z$, $\ell_{sm}$ & \textit{742.99$\pm$2.17}&\textit{3403.00$\pm$3.62} \\
     & CoTTA & 1150.16$\pm$5.69&4134.16$\pm$9.15 \\   
     & ProxyTTA (Ours) & \textbf{724.77$\pm$5.18}&\textbf{3349.21$\pm$29.00} \\
    \toprule
\end{tabular}
    }
\caption{\textit{Additional results for test-time adaptation for depth completion on Waymo $\rightarrow$ VKITTI-FOG.}% \textbf{Bold} denotes best and \textit{italics} runner-up.
}
\vspace{-3.5mm}
\label{tab:experiments:ablation:waymovkitti}
\end{table}

\section*{G. Ablation study}\label{ablation}
Here, we ablate the effect of each loss term denoted with the checkmarks in \tabref{tab:experiments:ablation}.
Using \textit{sparse depth consistency loss} $\ell_z$ (Eqn. 4) alone can improve the pretrained model as it learns the shapes of the test domain.
However, because of the sparsity, the supervision signal is weak, leading the model to exhibit artifacts and distortions in the depth map. 
Including a \textit{local smoothness loss} $\ell_{sm}$ (Eqn. 5) mitigates this by propagating depth to nearby regions. 
However, without knowledge of 3D shapes compatible with the sparse points, the wrong predictions are sometimes propagated as in the left bounding box region from Row 1, Column 4 of Fig. 4.
The best-performing method employs the proposed proxy embeddings as a regularizer to guide the adaptation layer update.
As the proxy mapping produces test-time features that follow the distribution of the source domain, minimizing our \textit{proxy consistency} loss (Eqn. 6) implicitly aligns the test domain features to those of the source domain that are compatible with the 3D scene observed by the test-time sparse point cloud.
Not only does this improve overall performance, but it also reduces standard deviation in error, which can be interpreted as an increase in the stability of the adaptation. We show qualitative comparisons against BN Adapt in Fig. 4, where boxes highlight improvements by fixing erroneous propagation by local smoothness (e.g., bleeding effect, which is not mitigated by using image gradients as guidance in Eqn. 5).
Quantitatively, we improve over the baseline by an average of 21.09\% across all methods and datasets, demonstrating the efficacy of our proxy embedding.

\section*{H. KITTI $\rightarrow$ VKITTI results}
\label{KITTItoVKITTI}

Here, we present additional results on KITTI $\rightarrow$ VKITTI adaptation. Test-time adaptation results are shown in \tabref{tab:experiments:ablation:vkitti}. 
Consistent with the trends observed in the main paper, our method outperforms over both BN Adapt and CoTTA, with a 21.82\% improvement compared to BN Adapt and 12.6\% improvement over CoTTA.


\section*{I. Experiment with different source dataset}
\label{source-waymo}
In our main paper, the only source dataset for outdoor adaptation scenario was KITTI which is the most popular outdoor depth completion dataset.
To validate our method's applicability to models trained on diverse source datasets, we include additional results from adaptation scenarios using a model trained on the Waymo dataset, as shown in Table \ref{tab:experiments:ablation:waymovkitti}.
Our method shows an improvement over CoTTA and BN Adapt by 21.70\%.

A noteworthy observation from the Waymo adaptation results, when compared to the KITTI $\rightarrow$ VKITTI-fog results from the main paper, is that the adaptation result of KITTI outperforms that of Waymo. This difference is caused by from the domain discrepancies between KITTI and VKITTI-fog datasets versus the domain gap between Waymo and VKITTI-fog. For example, VKITTI's object appearances and resolution (1226$\times$370 for KITTI, and 1242$\times$375 for VKITTI) are more akin to those in the KITTI dataset.
Conversely, the Waymo dataset features higher resolution (1920$\times$1280) and different object shapes compared to KITTI and VKITTI.
Hence, the adaptation result is influenced by the extent of domain discrepancy between the source and target datasets.

\section*{J. Quantitative preliminary results}
To provide a precise observation, we provide the quantitative results of model sensitive study in Tab. \ref{table:exp:preliminary}.

\end{document}